%% file: FDVS-TCSVT.tex
\documentclass[lettersize,journal]{IEEEtran}
\usepackage{amsmath,amsfonts}
\usepackage{algorithmic}
\usepackage{algorithm}
\usepackage{array}
\usepackage[caption=false,font=normalsize,labelfont=sf,textfont=sf]{subfig}
\usepackage{textcomp}
\usepackage{stfloats}
\usepackage{url}
\usepackage{verbatim}
\usepackage{graphicx}
\usepackage{cite}
\usepackage{amssymb}
\usepackage{booktabs}
\usepackage{enumitem}
\usepackage{paralist}
\usepackage{colortbl}
\usepackage{multicol}
\usepackage{multirow}
\usepackage{subcaption}
\usepackage{makecell}
\usepackage{rotating}
\graphicspath{{imgs/}} 
\def\eg{\emph{e.g}.} 
\def\ie{\emph{i.e}.}

\def\Mat#1{{\boldsymbol{#1}}}

\usepackage{xspace}
\def\sexyname{FDVS\xspace}
\def\longname{Fine-Detailed Video Story generation\xspace}
\def\hierarchical{Bottom-up Video Interpretation Mechanism\xspace}

\include{notation}

\def\mytitle{Towards Long Video Understanding via Fine-detailed Video Story Generation}

\begin{document}

\title{\mytitle}

\author{Zeng You, Zhiquan Wen, Yaofo Chen, Xin Li, Runhao Zeng, Yaowei Wang, Mingkui Tan

\thanks{This work was supported in part by the National Natural Science Foundation of China (NSFC) under Grant 62072190; in part by the Key-Area Research and Development Program Guangdong Province under Grant 2018B010107001; in part by the National Natural Science Foundation of China (NSFC) under Grant 62202311; in part by the Major Key Project of Peng Cheng Laboratory (PCL) PCL2023A08; in part by TCL Science and Technology Innovation Fund, Shenzhen Natural Science Foundation (the Stable Support Plan Program) under Grant 20220809180405001; and in part by the Excellent Science and Technology Creative Talent Training Program of Shenzhen Municipality under Grant RCBS20221008093224017.
}
\thanks{Zeng You is with the School of Future Technology, South China University of Technology, Guangzhou, China and also with Peng Cheng Laboratory, Shenzhen, China (e-mail: zengyou.yz@gmail.com).}
\thanks{Zhiquan Wen and Yaofo Chen are with the School of Software Engineering, South China University of Technology, Guangzhou, China and also with Peng Cheng Laboratory, Shenzhen, China (e-mail: sewenzhiquan@gmail.com, chenyaofo@gmail.com).}
\thanks{Xin Li is with Peng Cheng Laboratory, Shenzhen, China (e-mail: lix07@pcl.ac.cn).}
\thanks{Runhao Zeng is with the Artificial Intelligence Research Institute, Shenzhen MSU-BIT University and also with College of Mechatronics and Control Engineering, Shenzhen University (e-mail: zengrh@smbu.edu.cn).}
\thanks{Yaowei Wang is with Peng Cheng Laboratory, Shenzhen, China, and also with the School of Computer Science and Technology, Harbin Institute of Technology, Shenzhen, China (e-mail: wangyw@pcl.ac.cn).}
\thanks{Mingkui Tan is with the School of Software Engineering, South China University of Technology, Guangzhou 510006, China (e-mail:mingkuitan@scut.edu.cn).}
\thanks{Zeng You and Mingkui Tan are equal contributors.}
\thanks{Mingkui Tan, Yaowei Wang, and Runhao Zeng are corresponding authors.}}

\markboth{IEEE TRANSACTIONS ON CIRCUITS AND SYSTEMS FOR VIDEO TECHNOLOGY}%
{Shell \MakeLowercase{\textit{et al.}}: A Sample Article Using IEEEtran.cls for IEEE Journals}

\maketitle

\begin{abstract}
Long video understanding has become a critical task in computer vision, driving advancements across numerous applications from surveillance to content retrieval. Existing video understanding methods suffer from two challenges when dealing with long video understanding: intricate long-context relationship modeling and interference from redundancy. To tackle these challenges, we introduce \longname (\sexyname), which interprets long videos into detailed textual representations. Specifically, to achieve fine-grained modeling of long-temporal content, we propose a \hierarchical that progressively interprets video content from clips to video. To avoid interference from redundant information in videos, we introduce a Semantic Redundancy Reduction mechanism that removes redundancy at both the visual and textual levels. Our method transforms long videos into hierarchical textual representations that contain multi-granularity information of the video. With these representations, \sexyname is applicable to various tasks without any fine-tuning. We evaluate the proposed method across eight datasets spanning three tasks. The performance demonstrates the effectiveness and versatility of our method.
\end{abstract}


\input{sections/1_intro}
\input{sections/2_related_work}
\input{sections/3_method}

\input{sections/4_experiment}
\input{sections/5_conclusion}

\bibliographystyle{IEEEtran}
\bibliography{example_paper}

\end{document}

%% file: notation.tex
\DeclareMathAlphabet\mathbfcal{OMS}{cmsy}{b}{n}

\def\0{{\bf 0}}
\def\1{{\bf 1}}

\def\bV{{\bf V}}

\date{\today}

%% file: sections/1_intro.tex

\section{Introduction}
\label{sec:intro}
%
Video understanding aims to comprehend, interpret, and extract meaningful information from video data, which involves a range of tasks aimed at replicating human-like understanding of visual and temporal information present in videos.
It has become an important task in computer vision with a wide range of applications, such as video surveillance~\cite{li2023intermediary, sun2023video}, video auditing~\cite{xu2023self}, and video retrieval \cite{zhu2023complementarity, momalrg}. 
%
%
%
Despite notable advancements, the majority \cite{videoclip,gcm} are tailored for short-duration videos (\eg, 5-30 seconds).
In contrast, real-world videos often involve complex events, spanning from a few minutes to several hours.
%

%
Most existing methods \cite{liu2024evcap,wang2023multi,liu2023video,chen2024multilevel} for long video understanding employ a pre-trained backbone for deep feature extraction and a task-specific head for prediction.
These methods have a significant drawback: the need for fine-tuning on extensive annotated data when adapting to a new task.
This hinders their practical scalability, as the collection and annotation of long videos in real-world scenarios is time-consuming and costly. 
Many researchers \cite{videochat,videollama,moviechat,video-chatgpt,VidIL} endeavor to leverage LLMs to achieve open-ended comprehension of the video content. 
However, these methods either require substantial alignment fine-tuning with video-text pairs or fail to achieve an accurate and detailed understanding of long videos. 
This primarily stems from two challenges in long video comprehension.
%
\textbf{First}, long videos exhibit highly intricate content and prolonged temporal relationships. Accurately understanding long videos demands intricate temporal reasoning as these videos often unfold multiple events over time. 
%
\textbf{Second}, videos inherently contain a significant amount of redundancy in the temporal dimension \cite{moviechat}, which may interfere with the understanding and analysis of video content and introduce additional computational overhead. 

\begin{figure}[t]
    \centering
    \includegraphics[width = \linewidth]{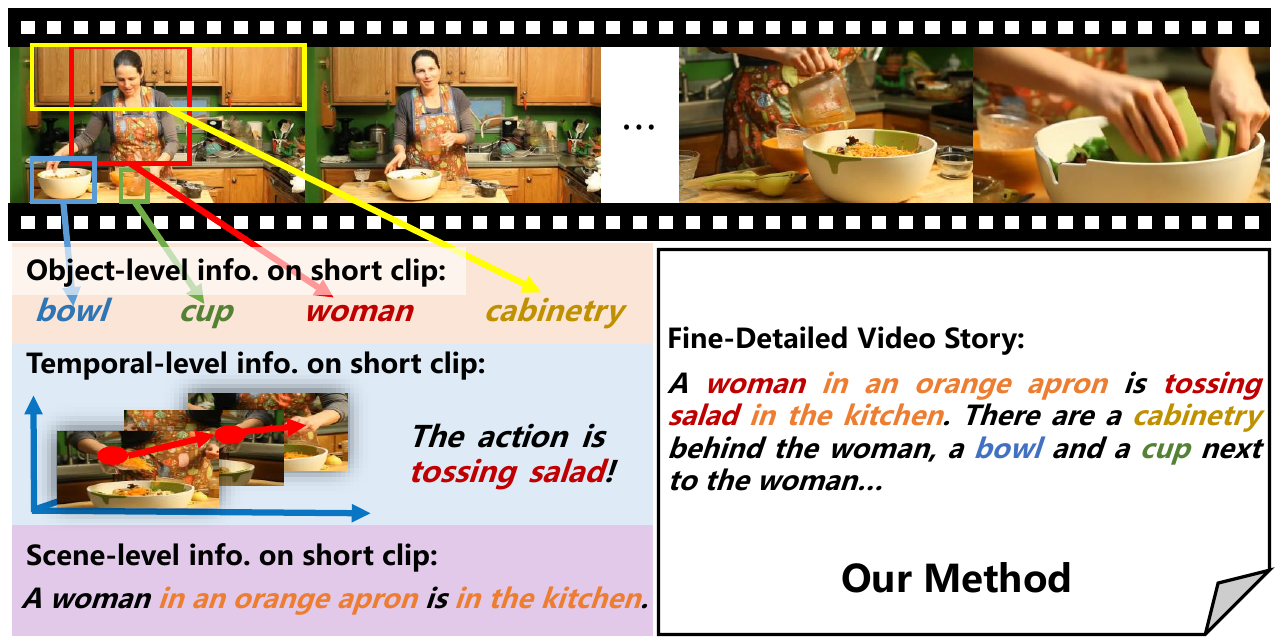}
    \caption{Illustration of video understanding with LLMs. To allow LLMs without visual perception to understand the video content, we provide perception information from three levels, \ie, object level, temporal level, and scene level.}
    \label{fig:cartoon}
\end{figure}

%
To achieve task adaptation across various tasks without requiring finetuning, we propose \longname (\sexyname), which can be adapted to various downstream tasks by representing videos as hierarchical textual information.
Specifically, we introduce a \hierarchical~aimed at simplifying the comprehension process for long-term video description generation. 
We break down videos into manageable clips based on keyframes.
Subsequently, we leverage well-trained vision foundation models as perception agents to comprehensively extract visual information from each frame within these clips. 
We then leverage Large Language Models (LLMs) to interpret the content of each short clip based on perception information, avoiding the bulk input of all information from the entire video at once. 
Finally, we instruct LLMs to summarize the video story with clip descriptions. 
This method progressively organizes and summarizes the fragmented information detected in the video frames into a structured story, comprehending complex video information at multiple granularities.
This diminishes the challenge of comprehending long video sequences and mitigating the issue of essential information being obscured. 

In addition, we propose a semantic redundancy reduction strategy to reduce the redundancy from both the visual level and the textual level to tackle the temporal redundancy issue. 
For visual level redundancy elimination, we determine and reduce the redundant frames within a clip via visual feature similarities. 
To address situations where visual pixels change due to factors like camera movement without altering content, we further employ higher-level textual information to eliminate redundant clips. 
It's worth noting that our extracted hierarchical textual representations contain information about the video at multiple scales, which are interpretable for both humans and machines.
We further explore a novel video understanding framework by applying our hierarchical textual representations rather than deep features across various downstream tasks without any cross-modal training or task-specific finetuning. 

Our contributions can be summarized as follows:
\begin{compactitem}
    \item We propose a long video understanding method named \longname~(\sexyname), which can adapt across various tasks without any fine-tuning or reliance on specific datasets.
    \item We develop a bottom-up video interpretation mechanism that processes videos to multiple granularities of textual information, facilitating comprehensive understanding of long videos across various levels.  
    \item We introduce a semantic redundancy reduction strategy to reduce the redundancy from both the visual level and textual level, eliminating the interference from redundancy.
\end{compactitem}




%% file: sections/2_related_work.tex
\section{Related Works}
\label{sec:related_works}
\subsection{Long-form Video Understanding}
While video understanding methods have made significant progress in tasks such as action recognition\cite{mou2023compressed,zheng2023dynamic,ZhaoLG022,slowfast,weng2020temporal,wang2023truncate}, video retrieval\cite{zhu2023complementarity,sun2023video,dong2022reading,ce,de}, and temporal action localization\cite{cao2022locvtp,gcm,winner-takes-all,yang2023cross,sun2022slow}, they are primarily designed for short videos. Real-world videos, however, often last from minutes to hours, consisting of multiple events. Long-form video understanding poses particular challenges due to the complexity of content and the spatio-temporal high dimensionality of the input. Methods for understanding long videos need to balance performance with efficiency. The methods can be grouped by long-range dependency modeling and sparsity methods.

long-range dependency modeling methods aim to enhance the long-temporal modeling capability. Several methods propose to achieve this through attention mechanisms and graphs. For instance, LF-VILA \cite{lf-vila} proposes a Temporal Window Attention (HTWA) mechanism to model long temporal dependency in long videos. VideoGraph \cite{videograph} proposes to learn an undirected graph from the dataset, where graph nodes represent the key concepts and edges represent the relationships between these key concepts. SVAG \cite{svag} proposes Supervoxel Attention Graphs for long-range dependency modeling. The
nodes of supervoxel attention graph are semantic supervoxels consisting of objects and motion cues in the video, while the edges are the spatiotemporal relations and feature similarity. Recently, several methods \cite{vis4mer,s4nd,s5} introduce Structured State-Space Sequence layers proposed in \cite{s4} to capture long temporal dependencies.

Sparsity methods aim to reduce the computational cost of long-form video understanding. Sparse sampling methods achieve this by sub-sampling meaningful frames from the video. These methods select frames based on saliency \cite{mgsampler}, adaptability \cite{adaframe}, or multimodal guidance \cite{barrios2023localizing}. Since sparse sampling unavoidably results in information loss, some methods \cite{mad, moviechat} aim to compress video content using a memory mechanism. 

Traditional approaches to video understanding often rely on pretrained backbone networks for deep feature extraction, followed by task-specific heads to make predictions. While effective, these methods require fine-tuning on annotated datasets for each new task, limiting their adaptability and generalization. Additionally, these approaches typically handle redundancy in video content by employing sparse sampling strategies or attention mechanisms that target visual redundancy alone, often neglecting cases where the visual content changes but the semantic meaning remains unchanged.In contrast to previous approaches, we employ textual representation to compress video content from the bottom up hierarchically. The extracted hierarchical textual representations can be applied to various video understanding tasks that require multi-grained information. Moreover, we introduce a semantic redundancy reduction strategy that tackles redundancy at both the visual and textual levels.

\subsection{Video Foundation Models}
The prevailing paradigm in building a video foundation model entails the initial pretraining of the model on an extensive large-scale video (or video-text) dataset, followed by fine-tuning it for specific downstream tasks~\cite{miech2020end,zellers2021merlot,hu2022scaling,yang2024gbc, zhang2022unsupervised}.
This approach has demonstrated its effectiveness across various applications, including video retrieval~\cite{videoclip,clip4clip} and human action recognition~\cite{actionRecSurvey, ZhaoLG022, pareek2021survey}.
The success of this approach relies on end-to-end training, utilizing pretext pretraining tasks such as masked language modeling~\cite{li2023lavender}, masked video modeling~\cite{videomae,videomaev2}, video-language masked modeling~\cite{violet}, video-text matching~\cite{wang2023all}, and video-text contrastive learning~\cite{internvideo,videoclip}.
However, a challenge arises when dealing with language-related tasks, such as video question answering and video captioning. These tasks require high-quality video-text pairs with detailed textual annotations, which are often lacking in the pretraining data.
In this paper, we enhance the capability for language-related tasks by leveraging a large language model as the ``brain" to organize the semantic information extracted from visual models.


\subsection{Video Understanding based on Large Language Models}
Recent advancements in large language models ~\cite{brown2020language,glm,llama} have significantly enhanced the capabilities of language generation. This progress have introduced innovative zero-shot capabilities for handling complex visual tasks, such as video question answering ~\cite{mist,jffusion,qa-dataset,anet-qa} and video captioning ~\cite{LinLL0G0LW22,vid2seq,anetcap,clip4clip}.
For instance, BLIP2~\cite{blip2} and MiniGPT-4 ~\cite{zhu2023minigpt} exhibit a plethora of advanced features enabling them to execute a variety of visual tasks in a zero-shot manner. This is achieved by coupling a frozen visual encoder with a frozen LLM, integrated through a trainable projection layer. Furthermore, video-language bridging methods~\cite{videochat,video-chatgpt,moviechat,videollama} expand into the realm of video understanding by fine-tuning the model with video-text paired data.
However, a limitation arises with these methodologies when they uniformly sample an identical number of frames from videos of varying lengths, potentially omitting crucial information from long-form videos. Our method addresses this by implementing a keyframe-based frame-sampling strategy that ensures the preservation of essential information despite the sparse sampling approach.
Besides, some recent works ~\cite{VidIL,internvid} translate video visuals into textual descriptions using perception tools such as object detectors and action recognizers before this data is processed by LLMs for further understanding. While existing methods may be inefficient due to their consideration of all visual content, our Bottom-up Video Interpretation Mechanism interprets long videos progressively, starting from frame-level information, moving to clip-level content, and finally summarizing the video story. This approach reduces the complexity of understanding long videos. In addition, our approach targets the selective removal of redundancy on both visual and semantic content. This targeted reduction results in more efficient processing, enhancing the overall performance of the system.

%% file: sections/3_method.tex
\begin{figure*}[t]
    \centering
    \includegraphics[width = \linewidth]{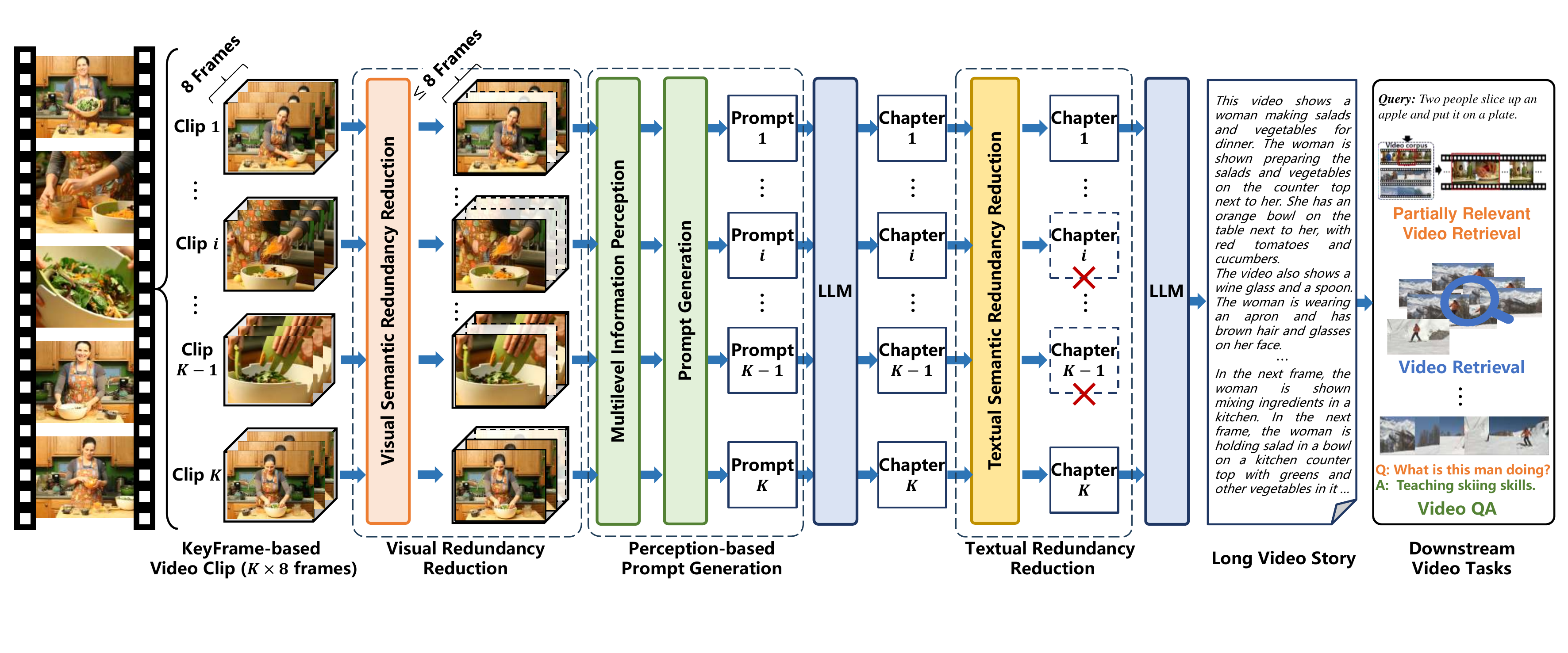}
    \caption{General pipeline of our method. We extract a compact hierarchical textual representation rather than deep features for downstream video understanding tasks. Given any video $\Mat{V}$, we initially segment and sample it into clips based on keyframes. Redundant frames within each clip are removed using a Visual Redundancy Reduction strategy. Subsequently, we employ three perception foundation models to extract visual information. An LLM describes the clip content using the perception information. Redundant clips are removed via Textual Redundancy Reduction. Finally, LLM summarizes the video story with the remaining chapters.}
    \label{fig:framework}
\end{figure*}

\begin{algorithm}[t]
    \renewcommand{\algorithmicrequire}{\textbf{Input:}}
    \renewcommand{\algorithmicensure}{\textbf{Output:}}
    \caption{General scheme of \sexyname}\label{alg:pipeline}
    \begin{algorithmic}[1]
    \REQUIRE  An untrimmed video $\bV$, pre-trained atomic agents $f$, pre-trained LLM $f_\text{LLM}$.\\
    \emph{// Video Segmentation.}
    \STATE Extract the key frames $\mathcal{X}_K$ in video $\bV$.
    \STATE Segment $\bV$ into clips $\{v_k\}_{k=1}^K$ based on $\mathcal{X}_K$.\\
    \FOR{$v_k$ in $\{v_k\}_{k=1}^K$}
    \STATE Uniformly sample frames from $v_i$ via Eq.(\ref{eq:sampling}).\\
    \STATE Reduce the redundant frames in $v_k$ via Visual-level Redundancy Elimination in Section.\ref{sec:visual_reduction}.\\
    \emph{// Extract perception information $\mathcal{A}$ for clip $v_k$}.\\
    \FOR{$f_n(\cdot)$ in $\mathcal{F}_a$}
    \STATE $a_k^n\leftarrow f_n(v_k)$.\\
    \ENDFOR\\
    \STATE $\mathcal{A}_k\leftarrow \{a_k^n\}_{n=1}^N$.\\
    \emph{// Summarize clip information $\mathcal{C}$}.\\
    \STATE $c_k\leftarrow f_\text{LLM}(\mathcal{A}_k)$.
    \ENDFOR
    \STATE $\mathcal{C} \leftarrow \left\{c_k\right\}_{k=1}^K$ \\
    \STATE Reduce the redundant clips via Semantic-level Redundancy Elimination in Section.\ref{sec:textual_reduction}\\
    \emph{// Summarize video story $\mathcal{V}$ using $f_{llm}$.}
    \STATE $\mathcal{V}\leftarrow f_\text{LLM}(\mathcal{C})$\\
    \ENSURE Hierarchical information $\mathcal{A},\mathcal{C}$, and $\mathcal{V}$.
    \end{algorithmic}
\end{algorithm}

\section{Fine-Detailed Video Understanding}
\label{sec:method}

\subsection{Overview}
%
Given an untrimmed video $\Mat{V}=\left\{x_1,x_2,\cdots,x_T\right\}$ with $T$ frames, where $x_t$ denotes the $t$-th frame, we aim to represent $\Mat{V}$ as textual representations using existing pre-trained vision foundation models $f$ and LLMs $f_{\text{LLM}}$. 
%
%
With multi-grained textual representation, our method is able to adapt across various tasks without any finetuning.

To comprehend the long-context videos comprehensively and accurately, we propose a \hierarchical~to interpret videos from clips to video. 
Specifically, we segment the given video $\Mat{V}$ into $K$ clips based on keyframes (see Section~\ref{sec:kerframe}). 
For each clip, we reduce the redundant frames via our visual semantic redundancy reduction strategy (see Section~\ref{sec:visual_reduction}). 
Then, we employ well-trained foundation models to detect and textualize perception information $\mathcal{A}$ (\eg, objects and attributes) from the remaining frames.
With this perception information, we then employ a Large Language Model (LLM) to generate a textual representation of the clip (\ie, chapter $c_k$).
(see Section \ref{sec:prompt_generation}).
Next, we reduce redundant clip chapters with our textual semantic redundancy reduction strategy (see Section~\ref{sec:textual_reduction}).
Furthermore, we leverage the LLM to reason and summarize the remaining chapters into a story $\mathcal{V}$ for the whole video. 
Our hierarchical textual representations (\ie, chapters and story) contain multi-granularity information, which can be applied to various video understanding tasks, \eg, video retrieval and video question answering.
The general pipeline is depicted in Figure~\ref{fig:framework} and Algorithm \ref{alg:pipeline}.

\subsection{Efficient Keyframe-based Video Segmentation}
\label{sec:kerframe}
Our approach involves representing videos as textual representations.
These representations are meticulously constructed from a series of elements, such as the categories and locations of objects in the video. 
This concept draws inspiration from the human understanding of the world, \ie, perceiving and interpreting distinct elements with senses.

  
Previous methods \cite{videochat,video-chatgpt, internvid} employ either uniform sampling mechanisms or tools like PySceneDetect \cite{scenedetect}  for arbitrary-length video segmentation. 
However, these approaches suffer from the following drawbacks: potential fragmentation of scenes and events, loss of key information in uniform sampling (especially in long videos), and low efficiency due to the requirement to decode all frames. 
To address the above issues, we devise an efficient video segmentation strategy that leverages 
the keyframes of the video. 
Specifically, we first extract the keyframes, \ie, the intra frames, $\mathcal{X}_K=\{x_{k}\}_{k=1}^K$, where $\mathcal{X}_K$ are $K$ keyframes sampled from $\Mat{V}$.\footnote{Extrating keyframes can be easily implemented by existing video decoding library, \eg, decord \cite{decord}.}
Based on the keyframes, we divide the video into $K$ clips, each beginning with $x_{k}$, while ending with $x_{k+1}$. 

After obtaining the keyframe-based video clips, we uniformly sample a fixed number of frames (8 frames) following previous methods~\cite{winner-takes-all,LinLL0G0LW22}.
Formally, the video clip is sampled as follows:
\begin{equation}
\label{eq:sampling}
        v_k = \left\{x_t\mid t \in S_u(x_k, x_{k+1})\right\},
\end{equation}
where $S_u$ denotes the uniform sampling method.

In addition to using intra-frame for key frame, motion vectors are another potential option. Motion vectors are estimated during video compression based on the assumption of rigid motion between consecutive frames. However, these motion vectors may be inaccurate, particularly in complex or fast-moving scenes, which can lead to segmentation errors. For objects exhibiting non-rigid motion, such as people walking or animals in motion, the shape and appearance of these objects can change significantly between frames. Motion vectors, which typically represent pixel-level translations, often fail to capture such complex transformations, making them less suitable for detecting scene boundaries or key content changes.

In contrast, intra-frame segmentation is independent of inter-frame information, meaning each frame is fully decoded and encoded independently. This provides a more detailed and reliable representation of the visual content, especially when dealing with long videos where the goal is to capture meaningful transitions in scenes or events without being affected by compression artifacts or estimation errors inherent in motion vectors. Thus, our method leverages intra-frame to ensure the accurate identification of key transitions, enhancing the quality and reliability of the segmentation process.

\subsection{Visual Semantic Redundancy Reduction}\label{sec:visual_reduction}
Neighboring frames of a video often share the same environment and background, with only slight positional changes in foreground objects, resulting in visual redundancy.
We should remove these redundant information since video understanding focuses more on varying content.
Existing methods~\cite{videochat,video-chatgpt,videollama,internvid,VidIL} typically involve either uniformly sampling a fixed number of frames or employing a significantly reduced frame rate for videos of various lengths.
This methods only partially reduces redundant frames but also tends to discard critical information. 

To address this, we propose a visual variation-aware redundancy reduction strategy, which removes redundant frames with minor variations in visual appearance and retains frames with more dynamic information. Specifically, for each clip $v_k$ (as previously defined in Eqn.~(\ref{eq:sampling})), we use $x_{k}$ to denote the key frame and $x_{t}$ to denote the subsquent frames sampled in this clip. We extract the visual feature $h_t$ for each frame $x_t$ by $h_t = f_{\text{img}}(x_t)$, where $f_{\text{img}}$ is the pretrained image encoder, \eg, CLIP \cite{clip}. Then we compute the cosine distance $s_t$ between $x_t$ and key frame $x_{k}$  by
\begin{equation}
    s_t = \frac{h_{k}\cdot h_t}{\Vert h_{k}\Vert \cdot \Vert h_t \Vert},
\end{equation}
where $h_{k}\small{=}f_{\text{img}}(x_{k})$ is the feature of the key frame.
We remove the redundant frames in clip $v_k$ when their similarity distance is higher than the threshold $\bar{s}$, which is defined as the average of all similarity distances $s_t$ within this clip.

\subsection{Perception-based
Prompt Generation}\label{sec:prompt_generation}
\begin{figure}[t]
    \centering
    \includegraphics[width = \linewidth]{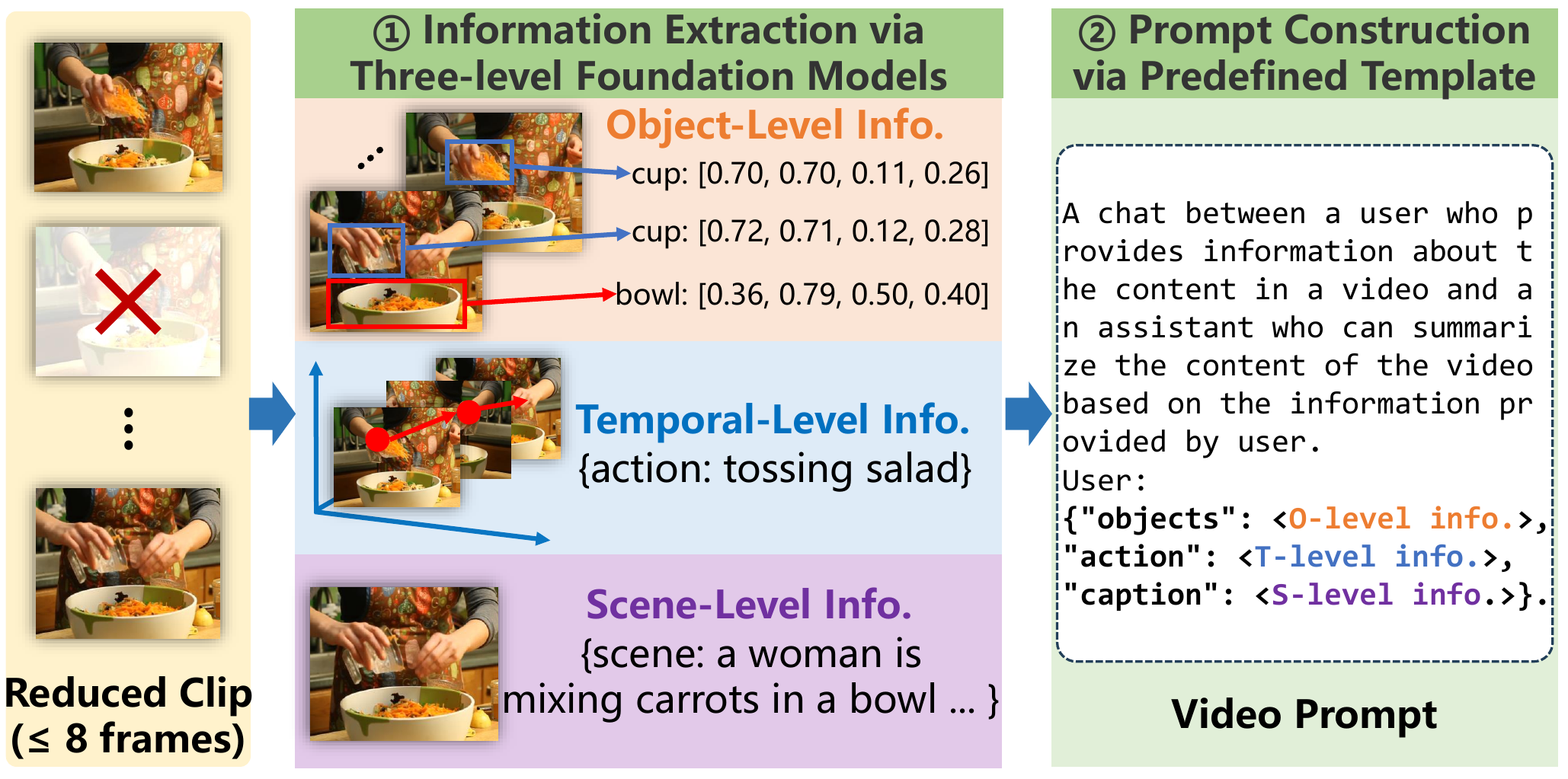}
    \caption{Illustration of information extraction via three-level agent and prompt organization via a predefined template. We leverage well-trained vision models as perception agents to comprehensively extract visual information from frames. Then, we leverage LLMs to interpret each clip's content based on perception information.}
    \label{fig:subfigure}
\end{figure}

Videos comprise a wealth of information, manifesting through both spatial and temporal dimensions. Spatial dimensions illustrate the objects with corresponding positioning and scene's content. Conversely, temporal dimensions sketch the progression of events, providing a narrative flow.
We employ three-level foundation models to capture information across specific dimensions (see Figure~\ref{fig:subfigure}). At the object level, we use an object detector (\eg, Grounding DINO~\cite{dino}) to identify categories and positions of objects within the image. At the temporal level, we deploy an action recognizer (\eg, InternVideo~\cite{internvid}) to classify the overall temporal actions observed in the footage. At the scene level, we apply an image captioner (\eg, BLIP2~\cite{blip2}) produces multiple sentences to describe the scene comprehensively.
Formally, We extract information $\mathcal{A}_k$ for the $k$-th clip $v_k$ by:
\begin{equation}
a_k^n = f_n(v_k),\quad \mathcal{A}_k = \left\{a_k^n\right\}_{n=1}^{N},
\end{equation}
where $f_n$ is the $n$-th perception model at different levels, $a_k^n$ is the information extracted by $f_n$ from the $k$-th clip.

Subsequently, we exploit LLMs to summarize the perception information $\{\mathcal{A}_k\}_{k=1}^{K}$, derived from a series of video clips $\{{v_k}\}_{k=1}^{K}$.
A naive approach is to directly input all the extracted perception information to LLMs. However, the direct summarization of all information $\{\mathcal{A}_k\}_{k=1}^{K}$ by LLMs would be challenging, primarily owing to the complex objects and long temporal relationships inherent in videos.
To mitigate this difficulty, our approach involves a hierarchical summarization process at both the clip and video level. Specifically, we arrange the perceptual information $\mathcal{A}_k$ of each clip using a predefined prompt template in Table \ref{tab:prompt_app}, maintaining their temporal sequence. This structured information enables the LLM to generate a concise chapter for each clip, denoted as $c_k$, based on its corresponding perceptual information $\mathcal{A}_k$.
Next, we compile these chapters into a set $\mathcal{C} = \{{c_i}\}_{i=1}^{L}$ and feed them into the LLM, which then constructs a cohesive narrative $\mathcal{V}$ for the entire video. The process can be formalized as:
\begin{equation}
c_k = f_{\text{LLM}}(\mathcal{A}_k),\quad \mathcal{C} = \left\{c_k\right\}_{k=1}^{L}, \quad \mathcal{V} = f_{\text{LLM}}(\mathcal{C}).
\end{equation}
It is important to note that before we feed the chapters $\mathcal{C}$ into the LLM, we would reduce textual redundancy among the chapters, which is further detailed in Section~\ref{sec:textual_reduction}.

\begin{figure}[t]
    \centering
    \includegraphics[width = \linewidth]{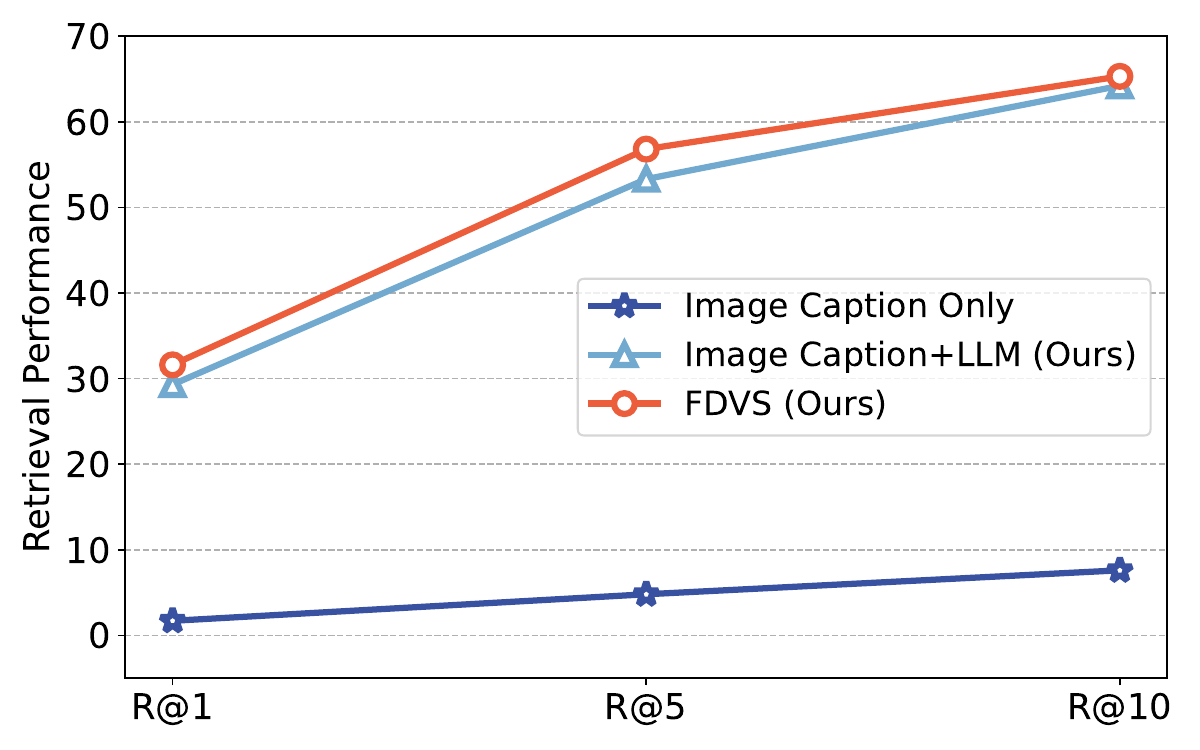}
    \caption{Comparison against image caption methods on MSRVTT over the zero-shot text-to-video retrieval task. 
    }
    \label{fig:zreo_shot_retrieval_msrvtt}
\end{figure}

Our \sexyname, which generates hierarchical textual representations of videos, supports a wide range of downstream tasks, such as video retrieval and video question answering.
However, relying solely on Image Caption models is insufficient for these purposes.
The "Image Caption Only" results in Figure~\ref{fig:zreo_shot_retrieval_msrvtt}, showcasing a mere 1.7\% R@1 for the text-to-video retrieval task.
The reason may be that the Image Caption model provides fragmented and potentially redundant information, leading to poor performance. 
We observe substantial improvements when we apply summarizing hierarchically with LLMs (29.3 R@1) highlighting the effectiveness of our method.
The culmination of employing our comprehensive three-level perception models, \ie, Full method (Ours), further enhances performance, achieving a notable 31.6 R@1.
This progression underscores the superiority and effectiveness of our \sexyname in advancing video understanding.

\subsection{Textual Semantic Redundancy Reduction}\label{sec:textual_reduction}

In real-world scenarios, the video shooting process is difficult to control, such as camera movement and zooming, which may bring about changes in appearance, but the semantic content is consistent. 
For example, we may shoot the same content from different angles, especially in a movie.
In other words, in the chapter set $\mathcal{C} \small{=} \left\{c_i\right\}_{i=1}^{L}$,
when a chapter $c_{i'}$ is similar as the previous ones, this clip $c_{i'}$ can be regarded as redundancy and should be reduced.

To this end, we introduce similarity measure $d_i\small{=}\text{sim}(h_i,\bar M)$ between $h_i$ and short-term historical information $\bar M$ to help determine whether we should remove the clip $v_i$.
Here, $h_i \small{=} f_{\text{txt}}(c_i)$ denotes the textual semantic features for chapter $c_i$, extracted using a pre-trained text encoder $f_{\text{txt}}$, \eg, Sentence-BERT~\cite{sentencebert}.
The historical average $\bar{M}$ is computed as $\bar{M}=\frac{1}{l}\sum_{j=i-l}^{i} h_{j}$, reflecting the average textual features of the preceding $l$ chapters.
In cases where obtaining $l$ chapters is not possible, we adjust to use as many as are available, ensuring non-zero historical context.
To avoid the selection of the similarity threshold, we remove a chapter if its similarity $d_i$ exceeds the mean similarity $\bar{d}$, calculated across all chapters.

%% file: sections/4_experiment.tex
\begin{table}[h]
\centering
\caption{The datasets used for evaluations.}
\label{tab:dataset}
\small
\resizebox{\linewidth}{!}{\begin{tabular}{c c} 
\hline
Task              & Datasets  \\ 
\hline
PRVR         & ActivityNet
Captions, Charades  \\
Video Retrieval   &   ActivityNet
Captions, MSRVTT\\
Video QA & MSRVTT-QA, ActivityNet-QA, EgoSchema, NExT-QA \\
\hline
\end{tabular}}
\end{table}

\begin{table*}[!h]
\centering
\caption{The prompt templates for image caption, clip description, video story summarization, video question answering, and short answer summarization.}
\label{tab:prompt_app}
\begin{tabular}{>{\centering\hspace{0pt}}m{0.087\linewidth}|>{\hspace{0pt}}m{0.85\linewidth}} 
\hline
                           & \multicolumn{1}{>{\centering\arraybackslash\hspace{0pt}}m{0.85\linewidth}}{Prompt Template}                                                                                                                                                                                                                                                                                                                                                                                                                                                                                                                                                                                                                                                                                                                                                                                                                                                                                                                                                                                                                                                                                                                                                                                                                                                                                                                                                                                                                                                                                                                                                   \\ 
\hline
Image Caption              & Describe this picture in as much detail as possible, including where this picture is located, what objects are there and what color they are. Answer:                                                                                                                                                                                                                                                                                                                                                                                                                                                                                                                                                                                                                                                                                                                                                                                                                                                                                                                                                                                                                                                                                                                                                                                                                                                                                                                                                                                                                                                                                         \\ 
\hline
Clip Description & \textit{A chat between a user who provides information about the content in a video and an assistant who can summarize the content of the video based on the information provided by the user. User: \{``action category": \textless{}clip action\textgreater{}, ``objects": \textless{}object\textgreater{}, ``caption": \textless{}image caption\textgreater{}\}. Assistant: }                                                                                                                                                             \\ 
\hline
Video Story      & \textit{A chat between a user who provides information about the content in a video and an assistant who can describe the content of the video in detail using `first', `then', `after that',  and `finnally' based on the information provided by the user. User: At the beginning of the video, \textless{}clip description1\textgreater{}. Early in the video, \textless{}clip description2\textgreater{}. Later in the video, \textless{}clip description3\textgreater{}. At the end of the video, \textless{}clip description4\textgreater{}. Assistant:}  \\
\hline
Video QA                   & \textless{}s\textgreater{}{[}INST]\par{} \textless{}\textless{}SYS\textgreater{}\textgreater{}\textbackslash{}n\par{}You are a helpful, respectful and honest assistant. Always answer as helpfully as possible. The User will provide information about the content in a video and ask you a question about this video. You should answer the question winthin 100 word based on the information provided by the user.\textbackslash{}n\par{}\textless{}\textless{}/SYS\textgreater{}\textgreater{}\textbackslash{}n\textbackslash{}n\par{} {[}/INST] \par{}Sure, I can help you with that! \textless{}/s\textgreater{} \par{}\textless{}s\textgreater{} {[}INST]\textless{}video info\textgreater{} Answer the following question within 100 word base on the infomation above: \textless{}question\textgreater{}{[}/INST] Short Answer in 100 word without any explaination:                                                                                                                                                                                                                                                                                                                                                                                                                                                                                                                                                                                                                                                                                                                                                               \\ 
\hline
Short Answer Summarization & \textless{}s\textgreater{}{[}INST] \par{}\textless{}\textless{}SYS\textgreater{}\textgreater{}\textbackslash{}n\par{}You are a helpful, respectful and honest assistant. Always answer as helpfully as possible. When User provides a question and a long answer to the question, you should summarize the long answer into 1 or 2 word.\textbackslash{}n\par{}\textless{}\textless{}/SYS\textgreater{}\textgreater{}\textbackslash{}n\textbackslash{}n \par{}{[}/INST] \par{}Sure, I can help you with that! \textless{}/s\textgreater{} \textless{}s\textgreater{} \par{}{[}INST]Question: what is the color of the pants of a person kneeling on one knee Long answer: The color of the pants of the person kneeling on one knee in the video is black.[/INST] Short Answer in 1 or 2 word: black\par{}{[}INST]Question: what is the person in black doing Long answer: The person in black is holding a soccer ball and standing next to the person playing soccer on the beach.[/INST] Short Answer in 1 or 2 word: beach soccer\par{}{[}INST]Question: is the person in white indoors Long answer: No, the person in white is not indoors. The video frames show them standing next to birds and other objects outdoors, with trees, grass, and a river in the background.[/INST] Short Answer in 1 or 2 word: no\par{}{[}INST]Question: is the person in white outdoors Long answer: Yes, the person in white is outdoors in all the frames.[/INST] Short Answer in 1 or 2 word: yes\par{}{[}INST]Question: \textless{}question\textgreater{} Long answer: \textless{}long answer\textgreater{}{[}/INST] Short Answer in 1 or 2 word:  \\
\hline
\end{tabular}
\end{table*}

\section{Experiments}
\label{sec:experiment}

%
We evaluate \sexyname on three tasks following the public settings: partially relevant video retrieval (PRVR)~\cite{prvr,jsg-prvr}, video retrieval~\cite{hit,clip4clip}, and video question answering ~\cite{video-chatgpt,moviechat} on the datasets specified in the Table~\ref{tab:dataset}. 
In addition, we conduct extensive ablation studies to verify the contribution of every component.

\subsection{Dataset Details}
\noindent\textbf{ActivityNet Captions} \cite{anetcap} was initially designed for the dense video captioning task. It includes 20,000 untrimmed videos of daily activities downloaded from YouTube. The average length of the videos in this dataset is the largest among PRVR datasets with 180 seconds. On average, each video has approximately 3.7 moments with corresponding sentence descriptions. The average length of each sentence is 13.48 words. For a fair comparison, we use the same data partition as in \cite{prvr, jsg-prvr} for PRVR. For video retrieval, we follow previous works\cite{hit,Gabeur0AS20,clip4clip,ZhangHS18,eclipse} evaluate paragraph-to-video retrieval on which we concatenate all the sentences of each video to a paragraph. We report the video retrieval results on the \textit{val1} split. 

\noindent\textbf{Charades-STA} \cite{charades-sta} is an extension of Charades \cite{charades} that incorporates sentence temporal annotations. It comprises 6,670 videos with 16,128 sentence descriptions. On average, each video has approximately 2.4 moments with corresponding sentence descriptions. We adhere to the data partition outlined in \cite{prvr} and evaluate our method on the test dataset.

\noindent\textbf{MSRVTT} \cite{msrvtt} is a well-known dataset for text-video retrieveal and video question answering. It is composed of 10,000 videos with durations that range from 10 to 32 seconds. Each video is annotated with 20 sentences by Amazon Mechanical Turks. We follow JSFusion \cite{jffusion} use the test data 'test 1k-A', which contains 1,000 video-text pairs, for evaluation of zero-shot retrieval performance.

\noindent\textbf{MSRVTT-QA} \cite{qa-dataset} is a video question answering dataset based on MSRVTT dataset \cite{msrvtt}. The question-answer pairs are generated from video descriptions using a question auto-generation tool \cite{ques-gener}. The generated question-answer pairs are composed of 243K open-ended questions. 

\noindent\textbf{ActivityNet-QA} \cite{anet-qa} is derived from ActivityNet dataset\cite{anetcap}. It is composed of 58,000 QA pairs and 5,800 videos. This dataset is focused on verifying the long-term spatio-temporal reasoning performance of the QA models.

\noindent\textbf{EgoSchema} \cite{ego_schema} is a new long-form video question-answering dataset introduced recently. It consists of 5,000 multiple-choice question-answer pairs, involving 250 hours of egocentric videos derived from Ego4D \cite{ego4d}. Each question in this dataset necessitates the selection of one correct answer from five options. We evaluate our method on the subset of 500 questions with ground truth answers released by the authors.

\noindent\textbf{NExT-QA} \cite{nextqa} is a video question answering benchmark designed to propel video understanding from mere description to deeper explanation of temporal actions. It comprises 5,440 videos accompanied by approximately 52K manually annotated question-answer pairs. These pairs are categorized into causal, temporal, and descriptive questions.

\subsection{Experimental Setup}


\noindent\textbf{Implementation Details}.
We employ GroundingDINO \cite{dino} with an open-vocabulary capability to detect objects present in the video. We set the box threshold as 0.4 and the text threshold as 0.25. The target categories for detection are defined as the intersection of the categories in COCO and Object365, encompassing a wide array of everyday objects. With the pre-trained BLIP2 \cite{blip2} as our image captioner, we generate textual descriptions for the video frames. We apply InternVideo \cite{internvideo} as the action recognizer to identify the action categories of Kinetics-400\cite{kinetics} in clips.
For clip descriptions and video narrative generation, we employ Vicuna-v1.5 \cite{vicuna}, trained by fine-tuning Llama 2 on user-shared conversations collected from ShareGPT \cite{sharegpt}.
%
We set the temperature as 0.7, the repetition penalty as 1.0, and the maximum number of tokens as 100. We use the cosine distance function in semantic redundancy reduction. 
We use the image encoder of CLIP \cite{clip} in visual redundancy reduction and Sentence-BERT \cite{sentencebert} as the text encoder in semantic redundancy reduction. We set the length $L$ of local memory used in semantic redundancy reduction as 35 based on the ablation study in Table \ref{tab:window_length}.

%
It is worth noting that the foundation models utilized in our method are highly flexible. These foundation models can be substituted or incorporated as required for different tasks. It can further enhance the video understanding capabilities of our approach when employing more robust foundation models.

\noindent\textbf{Details about applying long stories to downstream tasks.} 
Based on our extracted textual representation, we transform PRVR and video retrieval into a text-to-text retrieval task. We reconceptualize the text-to-video retrieval challenge as a text-to-text retrieval task. The match score is calculated between the query text and the generated descriptions for each video, including clip descriptions and the whole video story, rather than the original video.
This approach enables us to achieve multi-granularity retrieval through our multi-level textual representations. It is adaptable to different granularities of query text. The same methodology is employed for video retrieval, a higher-level retrieval task.
We employ AnglE \cite{angle}, obtained by fine-tuning LLaMa \cite{llama} on semantic textual similarity, to conduct the retrieval task. 

Moreover, our textual representations are rich in detail, paving the way for more complex video understanding tasks, such as video QA.
Concretely, we use the reconstructed representations as reference information for LLMs. However, recognizing that not all information is pertinent to the question, we initially identify relevant snippets and eliminate irrelevant ones by assessing the similarity between the question and the snippet descriptions. Subsequently, we input the clip description of the relevant clip and the story of the entire video into the LLMs as reference information, tasking them with answering questions based on this contextual input. We organize the question, selected clip descriptions, and video story according to the prompt template in Table \ref{tab:prompt_app}.

\noindent\textbf{Details about Prompt Templates}. In our framework, all models do not require any subsequent training and fine-tuning and are only used for inference. We will specify the prompts used in our proposed framework in this section. 
%
\subsubsection{Image Caption} In this paper, we employ BLIP2 \cite{blip2}, a multi-modal foundation model known for its robust zero-shot capabilities, as the image captioner. 
Each remaining video frame after visual semantic reduction, along with the corresponding prompt in Table~\ref{tab:prompt_app}, is input into an BLIP 2 to generate detailed scene descriptions.

\subsubsection{Clip Description} For each clip $v_k$, the perceptual information $A_k$ is organized using the corresponding template in Table \ref{tab:prompt_app}. The LLMs infer the content of $v_k$ based on $A_k$. The \textit{\textless clip action\textgreater}~denotes the action category within $v_k$, and \textit{\textless image caption\textgreater} in template is the remaining frames' image captions, organized as \textit{'frame \textless t\textgreater: \textless caption\textgreater'}, where \textit{\textless t\textgreater} is the index for frame. \textit{\textless object\textgreater} refers to the object categories, positions, and sizes detected after removing redundant frames. The image is divided into nine regions [top-left, top, top-right, left, center, bottom-left, bottom, bottom-right], and each object's position is determined by the center coordinate within these regions:
\begin{compactitem}
    \item \textbf{Top-left}: \(x < 0.33\), \(y < 0.33\)
    \item \textbf{Top}: \(0.33 \leq x < 0.66\), \(y < 0.33\)
    \item \textbf{Top-right}: \(x \geq 0.66\), \(y < 0.33\)
    \item \textbf{Left}: \(x < 0.33\), \(0.33 \leq y < 0.66\)
    \item \textbf{Center}: \(0.33 \leq x < 0.66\), \(0.33 \leq y < 0.66\)
    \item \textbf{Right}: \(x \geq 0.66\), \(0.33 \leq y < 0.66\)
    \item \textbf{Bottom-left}: \(x < 0.33\), \(y \geq 0.66\)
    \item \textbf{Bottom}: \(0.33 \leq x < 0.66\), \(y \geq 0.66\)
    \item \textbf{Bottom-right}: \(x \geq 0.66\), \(y \geq 0.66\)
\end{compactitem}
The size is categorized as [large, medium, small] based on the area occupied:
\begin{compactitem}
    \item \textbf{small}: \(area < 0.33\)
    \item \textbf{medium}: \( 0.33 \leq area < 0.66\)
    \item \textbf{large}: \(area \geq 0.66\)
\end{compactitem}

\subsubsection{Video Story} Chapters of all clips are organized using the prompt template in Table \ref{tab:prompt_app}. The total duration is divided into four ranges:[beginning, early, later, final], and clips are categorized based on their start positions. The \textit{\textless clip description1\textgreater} to \textit{\textless clip description4\textgreater} fields are filled with chapters from the corresponding time ranges in chronological order.

\subsubsection{Video QA} The video information and questions are organized using the template in Table \ref{tab:prompt_app}, with LLMs answering based on the hierarchical textual information. \textit{\textless video info\textgreater} includes the relevant clip chapters and the story, while \textit{\textless question\textgreater} is the question to be answered.

\subsubsection{Short Answer Summarization} Since the answers answered by LLMs tend to be more detailed, while the dataset gives only one to two words of ground truth. For exact matching, we use LLMs to further summarize the answers to 1 or 2 words based on the in-context learning capability. This prompt instructs LLMs to summarize long answers into 1-2 word short answers. \textit{\textless question\textgreater} is the question being answered, and \textit{\textless long answer\textgreater} is the detailed response that needs summarization.

\begin{table*}[tb]
    \centering
\caption{Results of PRVR on ActivityNet Captions and Charades-STA.}
\label{tab:prvr}
\resizebox{0.7\linewidth}{!}{\begin{tabular}{cc|cccccc} 
\hline
\multicolumn{2}{c|}{\multirow{2}{*}{Method}} & \multicolumn{3}{c}{ActivityNet Captions} & \multicolumn{3}{c}{Charades-STA}  \\
\multicolumn{2}{c|}{}                        & R@1  & R@5  & R@10                       & R@1 & R@5 & R@10                  \\ 
\hline
\multirow{6}{*}{Supervised} & CE \cite{ce}       & 5.5  & 19.1 & 29.9                       & 1.3 & 4.5 & 7.3                   \\
                            & DE \cite{de}    & 5.6  & 18.8 & 29.4                       & 1.5 & 5.7 & 9.5                   \\
                            & ReLoCLNet \cite{relocnet}   & 5.7  & 18.9 & 30.0                       & 1.2 & 5.4 & 10.0                  \\
                            & XML \cite{xml}   & 5.3  & 19.4 & 30.6                       & 1.6 & 6.0 & 10.1                  \\
                            & MS-SL \cite{prvr}  & 7.1  & 22.5 & 34.7                       & 1.8 & 7.1 & 11.8                  \\
                            & DL-DKD \cite{DLDKD} & 8.0  & 25.0 & 37.5                       & -   & -   & -                     \\
\hline
\multirow{4}{*}{Zero-shot}  & MovieChat \cite{moviechat}      & 6.7  & 20.0 & 29.1                       & 1.6 & 4.8 & 7.6                   \\
                            & VideoChat \cite{videochat}      & 7.7  & 19.8 & 27.5                       & 1.6 & 4.7 & 7.4                   \\
                            & VideoLlaVA \cite{videollava}      & 9.2  & 24.8 & 35.1                       & 1.2 & 4.1 & 7.0                   \\
                            \cline{2-8}
                            & \sexyname~(ours)           & \textbf{14.0} & \textbf{32.5} & \textbf{43.9}                       & \textbf{1.8} & \textbf{5.6} & \textbf{9.5}                   \\
\hline
\end{tabular}}
\end{table*}

\subsection{Evaluation on Partially Relevant Video Retrieval}
%
Partially Relevant Video Retrieval (PRVR) aims to retrieve the partially relevant untrimmed videos, which contain at least one internal moment relevant to the given query, from a large collection of untrimmed videos. 
It is a more practical and fine-grained video understanding task than video retrieval. 
Given the absence of published work on zero-shot PRVR, we benchmark our approach against supervised learning methods. 
Furthermore, we demonstrate the superiority of our generated text representations by comparing our method with other large language models using the same retrieval strategy.
Table~\ref{tab:prvr} shows the results.
Despite our method and the Video Language Model (VLM) performing the PRVR task in the zero-shot setting, we still achieve comparable or even superior performance compared to supervised learning methods. 
This underscores the feasibility and effectiveness of utilizing textual representations for fine-grained video understanding tasks. 
Notably, our method outperforms supervised learning methods by a substantial margin on ActivityNet Captions, which is a long-form video dataset in PRVR, showcasing the capability of our proposed framework to extract textual representations with comprehensive and accurate details. 

\begin{table*}[htb]
\centering
\begin{minipage}[t]{0.49\linewidth}
\centering
\caption{Exact matching accuracy (\%) of zero-shot video question answering on MSRVTT-QA and ActivityNet-QA. The best performance is denoted by \textbf{bold} numbers.}
\resizebox{\textwidth}{!}{\begin{tabular}{lc|cc} 
\hline
Method         & \#V-T Data & MSRVTT-QA & ActivityNet-QA  \\ 
\hline
Just Ask \cite{justask}      & 69M           & 2.9       & 12.2            \\
LAVENDER \cite{li2023lavender}      & 5M            & 4.5       & -               \\
MERLOT Reserve \cite{merlotreserve}& 1B            & 5.8       & -               \\
FrozenBiLM \cite{frozenbilm}    & 10M           & 6.4       & 16.7            \\
HiTeA \cite{hitea}         & 5M            & 8.6       & -               \\
\sexyname~(ours)          & \textbf{0M}            & \textbf{14.8}      & \textbf{21.2}            \\
\hline
\end{tabular}
}
\label{tab:exa_qa}
\end{minipage}
    \hfill
\begin{minipage}[t]{0.49\linewidth}
\centering
\caption{LLM-Assisted evaluation results of zero-shot video question answering on MSRVTT-QA and ActivityNet-QA. The best performance is denoted by \textbf{bold} numbers.}
\resizebox{\linewidth}{!}{
\begin{tabular}{cc|cc|cc} 
\hline
\multirow{2}{*}{Method} & \multirow{2}{*}{\#V-T Data} & \multicolumn{2}{c|}{MSRVTT-QA} & \multicolumn{2}{c}{ActivityNet-QA}  \\
                        &                                & Acc.(\%) & score               & Acc.(\%) & score                    \\ 
\hline
FrozenBiLM \cite{frozenbilm}              & 10M                            & 16.8     & -                   & 25.0     & -                        \\
VideoChat \cite{videochat}               & 27M                          & 45.0     & 2.3                 & 26.5     & 2.2                      \\
Video-ChatGPT \cite{video-chatgpt}              & 2M                            & 49.3     & 2.8                & 45.3     & 3.3                      \\
MovieChat \cite{moviechat}              & 410M                              & 49.7     & 2.9                 & 51.5     & 3.1                      \\
VideoChat2 \cite{videochat}              & 67M                              & 54.1     & 3.3                 & 49.1     & 3.3                      \\
LLaMA-VID  \cite{videollama}             & 2M                              & \textbf{58.9}     & 3.3                 & 47.5     & 3.3                      \\
\sexyname~(ours)                   & \textbf{0M}                             & 53.7     & \textbf{3.3}                 & \textbf{53.4}     & \textbf{3.4}                      \\
\hline
\end{tabular}
}
\label{tab:gpt_qa}
\end{minipage}
\vspace{-3mm}
\end{table*}

\subsection{Evaluation on Video Question Answering}
We further evaluate the video understanding capability on video questing answering tasks following \cite{moviechat,video-chatgpt}.
%
In addition to traditional exact matching, we also apply LLM-Assisted Evaluation for the video question-answering task following the methodology outlined in \cite{video-chatgpt, moviechat}.
%
%
Given the question, correct answer, and the predicted answer by the model, GPT-3.5 is expected to provide a True or False judgment along with a relative score ranging from 0 to 5. 
The exact matching accuracy and LLM-Assisted evaluation results are reported in Tables \ref{tab:exa_qa} and \ref{tab:gpt_qa}, respectively.

\begin{figure*}[t]
    \centering
    \includegraphics[width = \linewidth]{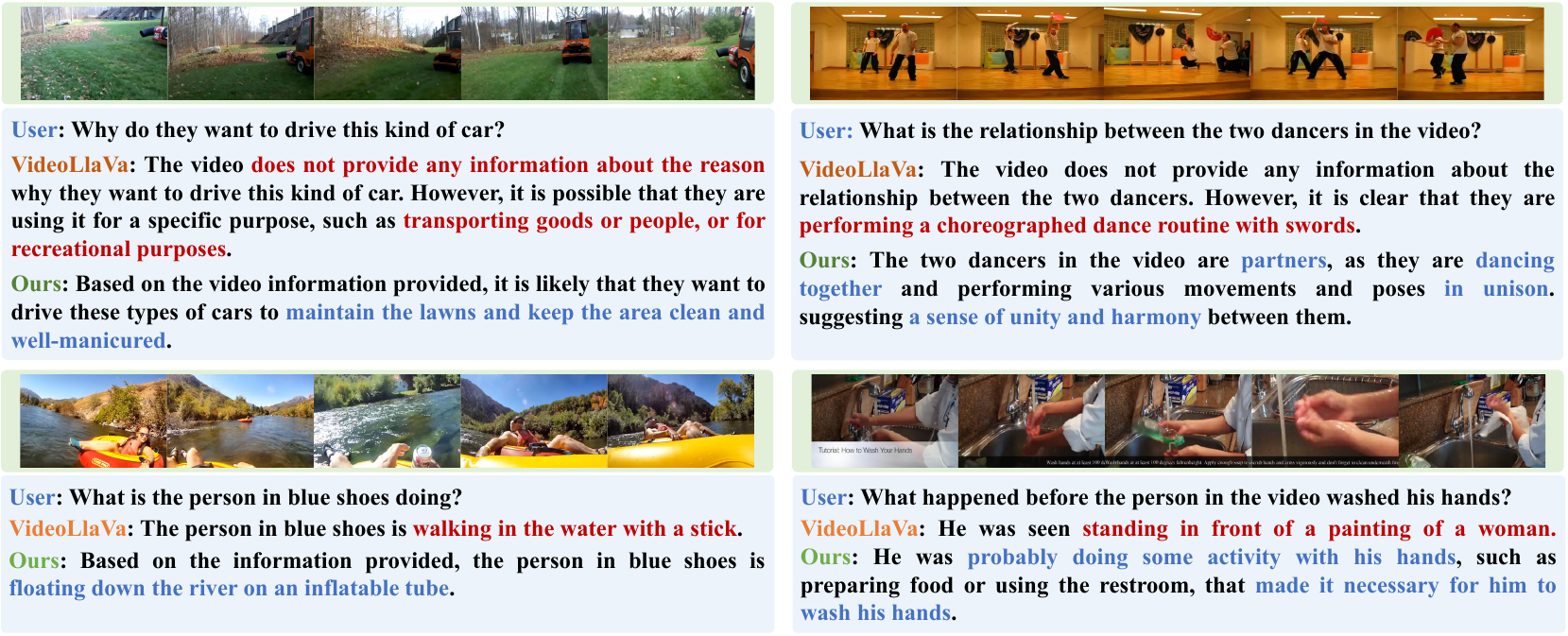}
    \caption{Qualitative results of our method and VideoLLaVa \cite{videollava}. The videos are from ActivityNet Captions.}
    \label{fig:qua_1}
\end{figure*}

In exact matching, we compare our method with the previous multi-modal foundation model, including Just Ask \cite{justask}, LAVENDER \cite{li2023lavender}, MERLOT Reserve \cite{merlotreserve}, FrozenBiLM \cite{frozenbilm}, and HiTeA \cite{hitea}. 
The results show that our approach achieves superior performance in terms of exact matching accuracy.

\begin{table}[b]
\centering
\caption{Multiple-choice QA accuracy on EgoSchema. The best performance is denoted by \textbf{bold} numbers.}
\label{tab:egochema}
\resizebox{0.6\linewidth}{!}{\begin{tabular}{lc} 
\hline
Method              & Acc.(\%)       \\ 
\hline
Random Choice       & 20.0           \\
SeViLA~\cite{yu2024self}              & 25.7           \\
Bard only (w/o Vision)~\cite{balazevic2024Memory}  & 27.0           \\
GPT-4 Turbo (w/o Vision)~\cite{balazevic2024Memory} & 31.0           \\
mPLUG-Owl~\cite{ye2023mplug}           & 33.8           \\
Bard + ImageViT~\cite{balazevic2024Memory}   & 35.0           \\
LLoVi~\cite{llovi}               & 40.4           \\
Bard + ShortViViT~\cite{balazevic2024Memory}            & 42.0           \\
\sexyname~(Ours)              & \textbf{42.4}  \\
\hline
\end{tabular}}
\end{table}

\begin{table}[t]
\centering
\caption{NExT-QA video question answering results. We report the accuracy for both the supervised and zero-shot methods.}
\label{tab:next-qa}
\begin{tabular}{c|cccc} 
\hline
Method     & Cau. (\%)     & Tem. (\%)     & Des. (\%)     & All(\%)        \\ 
\hline
SUPERVISED &               &               &               &                \\ 
\hline
ATP \cite{buch2022revisiting}        & 48.3          & 46.7          & 58.9          & 49.2           \\
Temp \cite{buch2022revisiting}       & 48.6          & 49.3          & 65.0          & 51.5           \\
TAATP \cite{xiao2022vgt}      & 53.1          & 50.2          & 66.8          & 54.3           \\
VGT \cite{xiao2022vgt}       & 52.3          & 55.1          & 64.1          & 55.0           \\
MIST \cite{mist}      & 54.6          & \textbf{56.6} & 66.9          & 57.1           \\
VFC \cite{momeni2023verbs}       & \textbf{57.6} & 53.3          & \textbf{72.8} & \textbf{58.6}  \\ 
\hline
ZERO-SHOT  &               &               &               &                \\ 
\hline
Flamingo \cite{alayrac2022flamingo}  & -             & -             & -             & 26.7           \\
CLIP \cite{clip}      & 43.6          & 38.1          & 57.0          & 43.9           \\
InternVid \cite{internvid} & 43.4          & 48.0          & 65.1          & 49.1           \\
VFC \cite{momeni2023verbs}       & 45.4          & \textbf{51.6} & 64.1          & 51.5           \\
\sexyname~(Ours)       & \textbf{58.8} & 51.3          & \textbf{69.7} & \textbf{58.1}  \\
\hline
\end{tabular}
\end{table}

In the LLM-Assisited evaluation, the comparison methods include VideoChat \cite{videochat}, MovieChat \cite{moviechat}, LLaMA-VID and VideoLLaVA \cite{videollava}. 
In comparison to previous large video-language models designed for video understanding, our framework also achieves superior performance in video question-answering, even without any training or fine-tuning on paired video-text data. 
To further showcase the effectiveness of our method in long-form video understanding, we present the question-answering accuracy on the EgoSchema \cite{ego_schema} and NExT-QA dataset \cite{nextqa} in Table \ref{tab:egochema} and Table \ref{tab:next-qa}. The comparison methods on EgoSchema consist of Bard, GPT-4, and LLoVi. Following previous works~\cite{balazevic2024Memory,wang2024videoagent}, we employ only the language modality of proprietary models. Even when compared to these large-scale proprietary models, our Fine-Detailed Video Stories (FDVS) method consistently achieves superior performance on long-form video QA. For example, our FDVS achieves 11.4\% improvements over GPT-4 Turbo without vision, and outperforms Bard with ImageViT by 9.4\%. Despite LLoVi demonstrating superior performance in long-form video understanding compared to Bard and GPT-4, it still falls short of our \sexyname~when utilizing the same large language model to comprehend video content. These results underscore the potency of our method in long-form video understanding. On NExt-QA, we compare our \sexyname with the supervised methods and zero-shot methods. Our method demonstrates the best performance among zero-shot approaches, achieving an overall average accuracy that surpasses the best competitor, VFC, by 6.4\%. Remarkably, our results are comparable to the supervised fine-tuned VFC, which underscores the strong capability of our method in long-form video understanding.

The success of our approach stems from the fact that, in contrast to other methods, our hierarchical textual representation can comprehensively extract and retain video information, spanning from details to high-level semantic information. 
Video question-answering tasks usually entail questions about specific details, an aspect where prior methods may lose crucial information or be affected by interference from redundant information. 
We also present some qualitative results in Fig.~\ref{fig:qua_1}. The visualized results show that our method achieves superior and accurate performance in content comprehension, causal reasoning, and relationship understanding. 

%
\begin{table}[tb]
    \centering
\caption{Results of zero-shot video retrieval on MSR-VTT. \textit{\# V-T Data} means the number of video-text paired data used to pre-train or finetune the model. The number \textbf{I} and \textbf{II} represent multimodal-based methods and LLM-based methods, respectively.}
\label{tab:retrieval on msrvtt}
\resizebox{\linewidth}{!}{\begin{tabular}{ccc|ccc} 
\hline
\multicolumn{2}{c}{\multirow{2}{*}{Method}}     & \multicolumn{1}{c}{\multirow{2}{*}{\# V-T Data}} & \multicolumn{3}{c}{MSRVTT}  \\
\multicolumn{2}{c}{}                            & \multicolumn{1}{c}{}                               & R@1  & R@5  & R@10          \\ 
\hline
\multirow{5}{*}{\textbf{I}} & VideoCLIP \cite{videoclip}  & 137M                                               & 10.4 & 22.2 & 30.0          \\
                                  & ALPRO \cite{alpro}      & 5M                                                 & 24.1 & 44.7 & 55.4          \\
                                  & VIOLET \cite{violet}     & 183M                                               & 25.9 & 49.5 & 59.7          \\
                                  & Singularity \cite{singularity} & 5M                                                 & 28.4 & 50.2 & 59.5          \\
                                  & HiTeA \cite{hitea}      & 5M                                                 & 29.9 & 54.2 & 62.9          \\
\hline
\multirow{4}{*}{\textbf{II}}        & VideoLLaVA \cite{videollava} & 2M                                               & 18.9 & 36.0 & 43.7          \\
                                  & MovieChat \cite{moviechat}  & 410M                                                  & 19.6 & 39.8 & 49.4          \\
                                  & VideoChat \cite{videochat}  & 27M                                              & 23.1 & 44.5 & 56.0          \\ 
\cline{2-6}
                                  & \sexyname~(ours)        & \textbf{0M}                                                 & \textbf{31.6} & \textbf{56.8} & \textbf{65.3}          \\
\hline
\end{tabular}}
\end{table}

\subsection{Evaluation on Video Retrieval}
We conduct a comparative evaluation of our method with two groups of approaches for zero-shot video retrieval. 
One group comprises multi-modal foundation models trained across video and text data, including VideoCLIP \cite{videoclip}. 
The other group involves LLM-based methods that extend Large Language Models (LLMs) to the video modal, such as VideoChat \cite{videochat} and MovieChat \cite{moviechat}.
For multi-modal foundation models, we directly test and report the performance of their models on the zero-shot video retrieval task. 
%
As large video language models cannot be directly used for video retrieval, we evaluate their performance using the same technical approach as ours. 
This involves using these models to generate a description for each video and then performing text retrieval. 
Our approach can also be seen as a way to extend the multi-modal capabilities of LLMs. 
The results on MSRVTT are presented in Table~\ref{tab:retrieval on msrvtt}. The results on ActivityNet Captions are shown in the Table~\ref{tab:retrieval on anet}.

Even though our method is not trained on any video-text pairwise data, it still achieves a remarkable performance on the video retrieval task. 
Our method outperforms all the other methods that draw on LLMs and even outperforms some pre-trained foundation models that are specialized for video retrieval tasks, such as VideoCLIP~\cite{videoclip}, ALPRO~\cite{alpro}, and VIOLET~\cite{violet}. 
Compared with previous LLM-based methods, our proposed approach achieves a significant performance gain with 7.2\% on R@1 metric on ActivityNet Captions. The ActivityNet Captions are considered as a long-form video dataset. The superior performance on ActivityNet Captions demonstrates the strong comprehension of our method for long-form videos. 
This demonstrates the high quality of the hierarchical textual representations generated by our method and the feasibility of using textual representations for the downstream video understanding tasks.

\begin{table}[t]
    \centering
    \centering
\caption{Results of zero-shot video retrieval on ActivityNet Captions. \textit{\#V-T Data} means the number of video-text paired data used to pre-train the model.}
\label{tab:retrieval on anet}
\resizebox{0.95\linewidth}{!}{\begin{tabular}{cc|ccc} 
\hline
\multirow{2}{*}{Method} & \multirow{2}{*}{\#V-T Data} & \multicolumn{3}{c}{ActivityNet Captions}         \\
                        &                                & R@1 & R@5  & R@10  \\ 
\hline
MovieChat~\cite{moviechat}& -                              & 14.9                     & 37.2 & 50.7  \\
VideoChat~\cite{videochat}               & 27M                            & 17.1                     & 36.3 & 46.6  \\ 
VideoLlaVa~\cite{videollava}              & 2M                             & 21.2                     & 46.1 & 58.5  \\
\hline
\sexyname (Ours)                    & \textbf{0M}                             & \textbf{28.4}                     & \textbf{57.5} & \textbf{71.1}  \\
\hline
\end{tabular}}
\end{table}

\begin{table}[b]
\centering
\caption{The average storage cost (bytes) of different methods for each video in MSRVTT.}\label{tab:storage}
\resizebox{\linewidth}{!}{\begin{tabular}{c|cccc} 
\hline
Method       & Clip4Clip & HiTeA & VideoClip & \sexyname~(ours)   \\ 
\hline
Avg. Storage & 840.0       & 168.0   & 280.0       & \textbf{87.3}  \\
\hline
\end{tabular}}
\end{table}

\subsection{Further Experiments and Discussions}

\noindent\textbf{Effect of storage saving.} 
Unlike previous methods such as \cite{clip4clip,hitea, videoclip}, which pre-extract deep features and store them for downstream tasks, our approach advocates the extraction and storage of hierarchical textual representations. This not only contributes to improved interpretability but also enhances storage efficiency for downstream tasks. 
We present the storage costs of features extracted by both previous methods and our textual representations in Table \ref{tab:storage}. Remarkably, the storage cost of our method is significantly lower than other approaches. Particularly noteworthy is the fact that, compared to Clip4Clip \cite{clip4clip}, our method achieves an impressive 89.6\% reduction in storage costs.

\noindent\textbf{Effect of different sampling strategies.} 
We propose an efficient keyframe-based sampling method that isolates video clips such that most frames within a clip belong to the same scene or event. This approach enhances the relevance and coherence of the sampled content. There are two alternative sampling methods: the uniform sampling method used in existing methods \cite{videochat,internvid, VidIL}, which segments videos into clips consisting of 8 frames without overlap, and the all-frame sampling method, which samples all frames from segmented clips and subsequently removes redundant frames using our visual semantic redundancy reduction strategy. We perform an ablation study to evaluate the effect of different sampling strategies. 
The results shown in Table \ref{tab:sampling} demonstrate the effectiveness of our sampling strategy on both performance and storage efficiency.

\begin{table}[t]
\centering
\caption{Zero-shot video retrieval results and storage cost (bytes) on MSRVTT using different sampling strategy. The default settings used in our framework are in the grey row.}\label{tab:sampling}
\resizebox{\linewidth}{!}{\begin{tabular}{c|cccc} 
\hline
Sample Strategy                                                  & R@1           & R@5           & R@10          & Storage Cost $\downarrow$  \\ 
\hline
All frames                                                       & 30.9          & 53.5          & 64.7          & 89.2                       \\
Uniform sampling                                                 & 30.9          & 51.4          & 62.6          & 319.8                      \\
\rowcolor[rgb]{0.867,0.867,0.867} Keyframe-based sampling & \textbf{31.6} & \textbf{56.8} & \textbf{65.3} & \textbf{87.3}              \\
\hline
\end{tabular}}
\end{table}

\noindent\textbf{Effect of redundant information reduction}. 
Videos often exhibit substantial redundancy in the temporal dimension. 
Firstly, the surrounding frames of a video frame typically feature the same environment, background, and objects, with only minor variations in the position or appearance of foreground objects, resulting in visual redundancy. 
Additionally, the video shooting process is often intricate, involving shot switching, camera movement, and zooming. While these actions may introduce visual changes, the underlying semantic information remains consistent, leading to semantic redundancy.

In this paper, we implement redundant information reduction at both the visual and semantic levels to mitigate computational costs and interference. 
The effectiveness of this redundant information reduction is demonstrated in Table \ref{tab:abl_rebundant}. 
The results indicate that both visual-level and semantic-level redundant reduction enhances the performance of our method in zero-shot video retrieval. 
This suggests that our proposed redundant information reduction method not only reduces unnecessary computational overhead but also contributes to the generation of more accurate textual descriptions. 

\begin{table}[b]
\centering
\caption{Ablation on Redundant Information Reduction. We evaluate the effects by conducting zero-shot text-to-video retrieval on ActivityNet Captions. Vis. Redu. denotes visual semantic redundancy reduction, while Text. Redu. represents textual semantic redundancy reduction The default settings used in our framework are in the grey row.}
\label{tab:abl_rebundant}
\resizebox{0.95\linewidth}{!}{\begin{tabular}{cc|ccc} 
\hline
Text. Redu.  & Vis. Redu.  & R@1           & R@5           & R@10           \\ 
\hline
             &              & 24.3          & 50.3          & 62.1           \\
  $\checkmark$            & & 26.2          & 54.2          & 68.3           \\
 &      $\checkmark$        & 26.9          & 54.6          & 68.2           \\
\rowcolor[rgb]{0.867,0.867,0.867}$\checkmark$ & $\checkmark$ & \textbf{28.4} & \textbf{57.5} & \textbf{71.1}  \\
\hline
\end{tabular}}
\end{table}

\begin{table}[t]
    \centering
\caption{Ablation on perception information by conducting zero-shot text-to-video retrieval on MSRVTT and ActivityNet Captions. Object, Action, and Caption represent the object detector, action recognizer, and image captioner, respectively. The default settings are in the grey row.} \label{tab:abl_agent}
\resizebox{\linewidth}{!}{\begin{tabular}{ccc|ccc|ccc} 
\hline
\multicolumn{3}{c|}{Perception Information}                                  & \multicolumn{3}{c|}{MSRVTT}                   & \multicolumn{3}{c}{ActivityNet}                \\ 
\hline
Object                                         & Action       & Caption      & R@1           & R@5           & R@10          & R@1           & R@5           & R@10           \\ 
\hline
$\checkmark$                                   &              &              & 1.5           & 5.8           & 10.1          & 0.4           & 2.5           & 4.1            \\
                                               & $\checkmark$ &              & 2.7           & 9.8           & 16.8          & 0.6           & 2.7           & 4.7            \\
                                               &              & $\checkmark$ & 30.2          & 54.4          & 65.1          & 24.4          & 51.5          & 65.5           \\
$\checkmark$                                   & $\checkmark$ &              & 5.1           & 14.9          & 24.4          & 3.0           & 11.4          & 20.8           \\
$\checkmark$                                   &              & $\checkmark$ & 30.7          & 54.5          & 64.1          & 26.1          & 53.1          & 67.6           \\
                                               & $\checkmark$ & $\checkmark$ & 30.9          & 56.7          & \textbf{66.1} & 27.4          & 54.8          & 68.7           \\
\rowcolor[rgb]{0.867,0.867,0.867} $\checkmark$ & $\checkmark$ & $\checkmark$ & \textbf{31.6} & \textbf{56.8} & 65.3          & \textbf{28.4} & \textbf{57.5} & \textbf{71.1}  \\
\hline
\end{tabular}}
\end{table}

\begin{table}[b]
\centering
\caption{Comparative analysis of zero-shot video retrieval on MSRVTT dataset using a standalone video captioner and the combination of image captioner and action recognizer.}
\label{tab:video_captioner}
\begin{tabular}{c|ccc} 
\hline
Perception Models                                                           & R@1           & R@5           & R@10           \\ 
\hline
 Video Captioner               & 29.2          & 52.4          & 63.9           \\ 
\hline
\begin{tabular}[c]{@{}c@{}}Image Captioner \\Action Recognizer\end{tabular}                                              & \textbf{31.6} & \textbf{56.8} & \textbf{65.3}  \\
\hline
\end{tabular}
\end{table}

\noindent\textbf{Effect of perception information.} In our framework, we exploit three perceptual models—an object detector, an action recognizer, and an image captioner—to capture different aspects of video content, specifically object (spatial), temporal (action), and scene-level information. We performed ablation studies on MSRVTT and ActivityNet Captions to evaluate the contribution of each perceptual model to the overall performance. The results are reported in Table \ref{tab:abl_agent}. 

Our observations indicate that the image captioner plays a pivotal role in hierarchical text feature generation, as it furnishes the majority of content information, encompassing objects, scenes, events, attributes, etc. 
The object detector has the least impact on textual representation generation since most object information is already covered in image captions. 
Furthermore, relying solely on the change in object position and size information in each frame makes it challenging for Large Language Models (LLMs) to infer the events occurring in the video.
Although the object detector and action recognizer contribute less individually, their combined usage is most effective when generating textual representations. 
This effectiveness arises from the complementary information provided by object information and action categories, which can augment image captions and rectify hallucinations that may occur in image captions.

An alternative approach is to use a video captioner instead of the combination of an image captioner and an action recognizer. To explore this, we conducted additional experiments using zero-shot video retrieval on the MSRVTT dataset. Specifically, we replaced the image captioner and action recognizer with a video captioner, employing VideoChat \cite{videochat} as the video captioner to generate captions for each video clip. The experimental results, presented in Table \ref{tab:video_captioner}, indicate that the performance of the combined image captioner and action recognizer significantly surpasses that of the standalone video captioner.
The superior performance of the combined approach can be attributed to the detailed and accurate information captured by image captions and action categories, which the video captioner alone struggled to match. While video captioners generally summarize both visual and temporal information, they often fail to capture the fine-grained details and dynamic aspects that are critical for comprehensive video understanding. The image captioner excels at frame-level detail extraction, while the action recognizer provides explicit temporal dynamics, together forming a richer representation of the video content.
Our method requires no training or fine-tuning on any data. Therefore, all perceptual models are plug-and-play in \sexyname, and it is flexible to select and add different models as needed for various tasks.


\begin{table}[t]
\centering
\caption{Ablation studies on ActivityNet Captions with different vision foundation models (CLIP-base vs. CLIP-large) and large language models. The default setting used in our framework is in the grey column. We use CLIP-base and Vicuna-7B as default models in our framework for their efficiency.}
\label{tab:foundation_models}
\begin{tabular}{c|ccc} 
\hline
Method                                      & R@1                  & R@5                  & R@10                  \\ 
\hline
\multicolumn{1}{l}{Visual Encoder}          & \multicolumn{1}{l}{} & \multicolumn{1}{l}{} & \multicolumn{1}{l}{}  \\ 
\hline
\rowcolor[rgb]{0.867,0.867,0.867} CLIP-base & 31.6                 & 56.8                 & 65.3                  \\
CLIP-large                                  & \textbf{33.3}        & \textbf{56.6}        & \textbf{65.9}         \\ 
\hline
\multicolumn{1}{l}{LLM}                     & \multicolumn{1}{l}{} & \multicolumn{1}{l}{} & \multicolumn{1}{l}{}  \\ 
\hline
\rowcolor[rgb]{0.867,0.867,0.867} Vicuna-7B & 31.6                 & 56.8                 & 65.3                  \\
Vicuna-13B                                  & 34.3                 & 59.6                 & 71.8                  \\
ChatGPT 3.5                                 & \textbf{35.4}        & \textbf{61.7}        & \textbf{75.0}         \\
\hline
\end{tabular}
\end{table}

\noindent\textbf{Effect of different foundation models}. In our proposed framework, the selection of foundation models is designed to be flexible, allowing for customization based on specific application requirements. For the visual semantic redundancy reduction component, we utilize a CLIP-base model as the image encoder to extract visual features from video frames. While for generating clip chapters and video stories, we employ Vicuna-7B.

To evaluate the impact of different models, we conducted ablation experiments using various image encoders and large language models (LLMs). The results of these experiments are presented in Table \ref{tab:foundation_models}. Notably, our findings indicate that utilizing larger variants of the CLIP model as the image encoder significantly enhances the performance of our visual semantic redundancy reduction mechanism. Similarly, when employing more powerful LLMs, such as Vicuna-13B and ChatGPT 3.5, we observe further improvements in framework performance. However, for efficiency reasons, our implementation utilizes smaller models. These experimental results underscore the flexibility of our approach, demonstrating that the choice of foundation models can be tailored to meet specific application needs, thereby maximizing effectiveness while maintaining operational efficiency.

\begin{figure}[t]
    \centering
    \includegraphics[width=\linewidth]{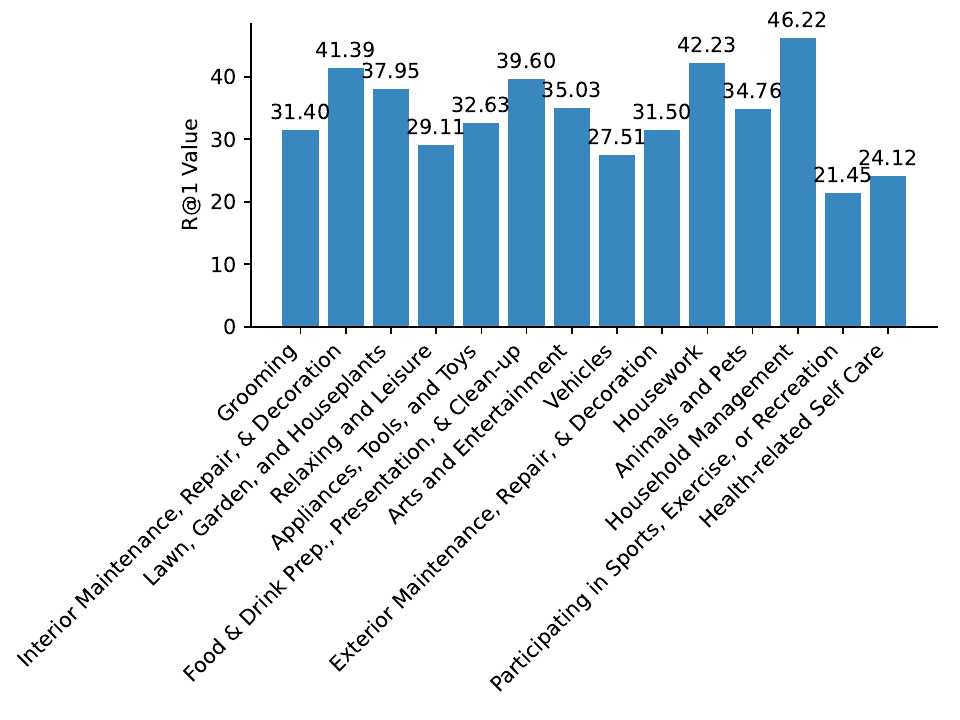}
    \caption{R@1 performance of \sexyname across various video categories on the ActivityNet dataset}
    \label{fig:result_cls}
\end{figure}

\noindent\textbf{Performance across various Video Categories}. We use multiple perception models to extract information from videos. Among these foundation models, GroundingDINO \cite{dino} and BLIP2 \cite{blip2} are open-set models and are applicable to a wide range of categories. Although InternVideo \cite{internvideo} is trained on fixed categories, it encompasses a majority of everyday scenarios. In addition, we leverage multiple-level perceptual models (object, scene, and temporal levels) allowing for a comprehensive understanding of various aspects of the videos. This multi-level perception ensures that even if InternVideo does not recognize certain actions, complementary information from other models can compensate, enhancing overall performance. To evaluate the performance of \sexyname on various categories, we report the R@1 performance of our method on different categories of videos from the ActivityNet dataset in Figure \ref{fig:result_cls}. The category labels are obtained from the second-level action category annotations of the action detection set of ActivityNet v1.3, ensuring that the category labels of the videos are accurate. The results illustrate that our method maintains balanced performance across these categories, with no significant performance drop observed in any specific category.

\begin{table}[t]
\centering
\caption{Comparison of inference speeds (FPS) between our method and other approaches during the Representation Extraction (Rep. Extra.) phase and downstream tasks (Video Retrieval, PRVR, and Video QA).}
\label{tab:infer_speed}
\begin{tabular}{c|cccc} 
\hline
Methods     & Rep. Extra.    & Video Retrieval & PRVR           & Video QA        \\ 
\hline
VideoChat   & \textbf{30.29} & 116.46          & 66.88          & 26.95           \\
VideoLlaVA  & 29.87          & 114.85          & 61.43          & 27.85           \\
MovieChat   & 20.72          & 102.36          & 60.91          & 22.55           \\
FDVS (Ours) & 13.22          & \textbf{125.75} & \textbf{80.76} & \textbf{37.73}  \\
\hline
\end{tabular}
\end{table}

\noindent\textbf{Effect of inference speed}. Our proposed \sexyname is designed specifically for processing offline long videos, where the primary goal is to first extract hierarchical textual representations from videos. These representations can then be directly utilized for various downstream understanding tasks without the need for additional fine-tuning on specific tasks or datasets.

To clarify the inference speed of our framework, we compare the inference speed of \sexyname with that of methods which also use LLMs.  We report the inference speed of our method compared with existing models in Table \ref{tab:infer_speed}, presented in terms of frames per second (FPS) during the representation extraction and downstream task stages. A key advantage of our approach lies in the efficiency gained in the downstream task phase. Once the hierarchical textual representations are extracted, they can be directly utilized for various downstream tasks such as video retrieval, PRVR, or question answering, without requiring any fine-tuning or further processing. Our \sexyname shows a significantly faster inference speed for downstream tasks compared to existing methods. The processing speed of our method is slower during the initial representation extraction phase compared to existing models. This is due to the complexity of converting the video content into detailed and hierarchical textual representations, which inherently involves multiple large models.

\begin{table}[t]
   \tiny
\caption{Ablation on the hierarchical textual representations in Video QA, \ie,  the number of clip descriptions $K$ and video story. We conduct the evaluation in zero-shot video question-answering with MSRVTT-QA and ActivityNet QA and report the exact matching accuracy. The default settings in our framework are in the grey row.}\label{tab:abl_vqa}
\resizebox{\linewidth}{!}{
\begin{tabular}{cc|cc} 
\hline
\multirow{2}{*}{$K$} & \multirow{2}{*}{Video Story} & \multicolumn{2}{c}{Accuracy}  \\
                                     &                              & MSRVTT-QA & ActivityNet-QA    \\ 
\hline
0                                    & $\checkmark$                 & 13.0      & 20.1              \\
1                                    & $\checkmark$                 & 14.0      & 21.3              \\
3                                    & $\checkmark$                 & 14.5      & 21.2              \\
\rowcolor[rgb]{0.867,0.867,0.867} 5                                    & $\checkmark$                 & \textbf{14.8}      & \textbf{21.2}              \\
5                                    &                              & 14.5      & 20.8              \\
7                                    & $\checkmark$                 & 14.8      & 21.0              \\
\hline
\end{tabular}}
\end{table}

\noindent\textbf{Effect of the hierarchical information used in video question-answering.} 
In video question-answering, we employed hierarchical textual representations, comprising clip descriptions and video stories. The video story encapsulates global information and the overall semantics of the video, while the clip description contains more detailed information. To assess the impact of hierarchical information in video question-answering, we conduct experiments and present the results in Table \ref{tab:abl_vqa}. Here, $K$ denotes the number of clip descriptions used in video question-answering. When $K=0$, we exclusively rely on the video story to furnish reference information for LLMs to answer the question. We observe that incorporating hierarchical information results in higher accuracy in video question-answering compared to relying solely on the video story. This improvement stems from the combined contributions of clip descriptions and the video story, which provide more accurate and comprehensive reference information.

\begin{table}[b]
\centering
\caption{Ablation Study on the impact of local memory length ($L$) in textual semantic redundancy elimination. This analysis evaluates the effectiveness of varying local memory lengths during video retrieval task using the ActivityNet Captions dataset. The default setting used in our framework is in the grey row.}
\label{tab:window_length}
\begin{tabular}{c|ccc} 
\hline
$L$                                  & R@1           & R@5           & R@10           \\ 
\hline
1                                    & 4.36          & 7.17          & 8.47           \\
11                                   & 27.1          & 51.1          & 68.0           \\
21                                   & 27.2          & 53.7          & 68.7           \\
31                                   & 27.3          & 56.2          & 69.4           \\
\rowcolor[rgb]{0.867,0.867,0.867} 35 & \textbf{28.4} & \textbf{57.5} & \textbf{71.1}  \\
41                                   & 27.3          & 56.8          & 69.3           \\
\hline
\end{tabular}
\end{table}

\noindent\textbf{Ablation study of local memory length $L$}. We remove textual semantic redundancy based on the historical average of local memory with length $L$. We conducted an ablation study on the length of local memory, and the results are presented in Table \ref{tab:window_length}. Here, \textit{SumR} denotes the sum of R@1, R@5, and R@10 values. It's noteworthy that a substantial performance degradation occurs when the length $L$ is too short, potentially due to the instability of the reference content. The optimal performance of our framework is achieved when $L=35$, and we adopt this value as the default setting.

\noindent\textbf{Ablation study on the number of frames sampled per clip}. We conduct ablation experiments regarding the sampling number for video clips, with results reported in Table \ref{tab: abl_num_frame}. The experiments indicate that a smaller sampling number (e.g., 4) results in performance degradation while increasing the sampling number to 8 yields improvements. However, further increases to 16 or 32 frames show only marginal performance gains. Thus, to balance computational load and performance, we have adopted a default sampling number of 8. This is also consistent with previous research \cite{winner-takes-all, LinLL0G0LW22} for fair comparisons.

\begin{table}[t]
\centering
\caption{The ablation study on the number of sampling frames in each segmented video clip. The default setting used in our framework is in the grey row.}
\label{tab: abl_num_frame}
\begin{tabular}{c|ccc} 
\hline
Num. Frames & R@1  & R@5  & R@10  \\ 
\hline
4           & 26.9 & 47.4 & 56.6  \\
\rowcolor[rgb]{0.867,0.867,0.867} 8           & 31.6 & 56.8 & 65.3  \\
16          & 31.7 & 57.4 & 66.0  \\
32          & 32.1 & 58.6 & 67.1  \\
\hline
\end{tabular}
\end{table}

\noindent\textbf{Discussion of advantages and potential limitations}. Our \sexyname is specifically designed for processing offline long videos. \sexyname, which extracts long video content into hierarchical textual representations, demonstrates the capability to be directly utilized for various downstream tasks without requiring any fine-tuning on task-specific datasets. This characteristic enables our method to exhibit strong zero-shot performance across multiple tasks, facilitating rapid adaptation to new applications. Furthermore, as shown in Table \ref{tab:infer_speed}, the inference speed for downstream tasks is notably high, providing a significant advantage in applications. Additionally, compared to traditional methods that extract deep features for downstream tasks, our textual representations incur lower storage overhead, as illustrated in Table \ref{tab:storage}.

However, to achieve a comprehensive understanding of long videos and to extract fine-
grained hierarchical textual representations, our approach utilizes perception models at three levels: object, scene, and temporal.
This multi-level perception enables us to capture different aspects of video content effectively. We also leverage large language
models (LLMs) for Bottom-up Video Interpretation and multi-granular representations, which enhances our representation
extraction process. This makes our FDVS slower during the representation extraction phase than existing methods.

\noindent\textbf{Discussion about optimization of \sexyname}. Our method is also scalable, allowing for easy improvements to the entire pipeline. For example, we can incorporate additional perception foundation models to enhance video perception, including a speech recognizer for audio perception. Additionally, since the use of large language models (LLMs) can be computationally intensive, optimizing the LLM's key-value (KV) cache can further improve computational efficiency.

%% file: sections/5_conclusion.tex
\section{Conclusion}
\label{sec:conclution}
In this paper, we introduce a comprehensive video understanding framework, referred to as \longname, that interprets videos as hierarchical textual representations, encompassing clip chapters and long video stories. To address the challenge of long-context video comprehension, we propose a \hierarchical~to represent videos with detailed information at multiple granularities, evolving from clip chapters to long video stories. To mitigate interference from inherent redundancy in videos, we propose a semantic redundancy reduction scheme aimed at eliminating redundancy at both the visual and textual levels. 
Leveraging the detailed hierarchical representations, our framework effortlessly adapts to various tasks without requiring specific fine-tuning.

%% file: FDVS-TCSVT.bbl
\begin{thebibliography}{100}
\providecommand{\url}[1]{#1}
\csname url@samestyle\endcsname
\providecommand{\newblock}{\relax}
\providecommand{\bibinfo}[2]{#2}
\providecommand{\BIBentrySTDinterwordspacing}{\spaceskip=0pt\relax}
\providecommand{\BIBentryALTinterwordstretchfactor}{4}
\providecommand{\BIBentryALTinterwordspacing}{\spaceskip=\fontdimen2\font plus
\BIBentryALTinterwordstretchfactor\fontdimen3\font minus
  \fontdimen4\font\relax}
\providecommand{\BIBforeignlanguage}[2]{{%
\expandafter\ifx\csname l@#1\endcsname\relax
\typeout{** WARNING: IEEEtran.bst: No hyphenation pattern has been}%
\typeout{** loaded for the language `#1'. Using the pattern for}%
\typeout{** the default language instead.}%
\else
\language=\csname l@#1\endcsname
\fi
#2}}
\providecommand{\BIBdecl}{\relax}
\BIBdecl

\bibitem{li2023intermediary}
H.~Li, M.~Liu, Z.~Hu, F.~Nie, and Z.~Yu, ``Intermediary-guided bidirectional
  spatial-temporal aggregation network for video-based visible-infrared person
  re-identification,'' \emph{IEEE Transactions on Circuits and Systems for
  Video Technology}, 2023.

\bibitem{sun2023video}
X.~Sun, J.~Gao, Y.~Zhu, X.~Wang, and X.~Zhou, ``Video moment retrieval via
  comprehensive relation-aware network,'' \emph{IEEE Transactions on Circuits
  and Systems for Video Technology}, 2023.

\bibitem{xu2023self}
Y.~Xu, X.~Li, L.~Pan, W.~Sang, P.~Wei, and L.~Zhu, ``Self-supervised
  adversarial video summarizer with context latent sequence learning,''
  \emph{IEEE Transactions on Circuits and Systems for Video Technology}, 2023.

\bibitem{zhu2023complementarity}
J.~Zhu, P.~Zeng, L.~Gao, G.~Li, D.~Liao, and J.~Song, ``Complementarity-aware
  space learning for video-text retrieval,'' \emph{IEEE Transactions on
  Circuits and Systems for Video Technology}, 2023.

\bibitem{momalrg}
Z.~Luo, Z.~Durante, L.~Li, W.~Xie, R.~Liu, E.~Jin, Z.~Huang, L.~Y. Li, J.~Wu,
  J.~C. Niebles \emph{et~al.}, ``Moma-lrg: Language-refined graphs for
  multi-object multi-actor activity parsing,'' \emph{Advances in Neural
  Information Processing Systems}, vol.~35, pp. 5282--5298, 2022.

\bibitem{videoclip}
H.~Xu, G.~Ghosh, P.-Y. Huang, D.~Okhonko, A.~Aghajanyan, F.~Metze,
  L.~Zettlemoyer, and C.~Feichtenhofer, ``Videoclip: Contrastive pre-training
  for zero-shot video-text understanding,'' in \emph{Proceedings of Conference
  on Empirical Methods in Natural Language Processing}, 2021, pp. 6787--6800.

\bibitem{gcm}
R.~Zeng, W.~Huang, M.~Tan, Y.~Rong, P.~Zhao, J.~Huang, and C.~Gan, ``Graph
  convolutional module for temporal action localization in videos,'' \emph{IEEE
  Transactions on Pattern Analysis and Machine Intelligence}, vol.~44, no.~10,
  pp. 6209--6223, 2022.

\bibitem{liu2024evcap}
S.~Liu, A.~Li, Y.~Zhao, J.~Wang, and Y.~Wang, ``Evcap: Element-aware video
  captioning,'' \emph{IEEE Transactions on Circuits and Systems for Video
  Technology}, 2024.

\bibitem{wang2023multi}
Y.~Wang, M.~Liu, J.~Wu, and L.~Nie, ``Multi-granularity interaction and
  integration network for video question answering,'' \emph{IEEE Transactions
  on Circuits and Systems for Video Technology}, vol.~33, no.~12, pp.
  7684--7695, 2023.

\bibitem{liu2023video}
J.~Liu, G.~Wang, J.~Xie, F.~Zhou, and H.~Xu, ``Video question answering with
  semantic disentanglement and reasoning,'' \emph{IEEE Transactions on Circuits
  and Systems for Video Technology}, 2023.

\bibitem{chen2024multilevel}
L.~Chen, Z.~Deng, L.~Liu, and S.~Yin, ``Multilevel semantic interaction
  alignment for video--text cross-modal retrieval,'' \emph{IEEE Transactions on
  Circuits and Systems for Video Technology}, 2024.

\bibitem{videochat}
K.~Li, Y.~He, Y.~Wang, Y.~Li, W.~Wang, P.~Luo, Y.~Wang, L.~Wang, and Y.~Qiao,
  ``Videochat: Chat-centric video understanding,'' \emph{arXiv preprint
  arXiv:2305.06355}, 2023.

\bibitem{videollama}
H.~Zhang, X.~Li, and L.~Bing, ``Video-llama: An instruction-tuned audio-visual
  language model for video understanding,'' in \emph{Proceedings of Conference
  on Empirical Methods in Natural Language Processing}, 2023, pp. 543--553.

\bibitem{moviechat}
E.~Song, W.~Chai, G.~Wang, Y.~Zhang, H.~Zhou, F.~Wu, X.~Guo, T.~Ye, Y.~Lu,
  J.-N. Hwang \emph{et~al.}, ``Moviechat: From dense token to sparse memory for
  long video understanding,'' \emph{arXiv preprint arXiv:2307.16449}, 2023.

\bibitem{video-chatgpt}
M.~Maaz, H.~Rasheed, S.~Khan, and F.~S. Khan, ``Video-chatgpt: Towards detailed
  video understanding via large vision and language models,'' \emph{arXiv
  preprint arXiv:2306.05424}, 2023.

\bibitem{VidIL}
Z.~Wang, M.~Li, R.~Xu, L.~Zhou, J.~Lei, X.~Lin, S.~Wang, Z.~Yang, C.~Zhu,
  D.~Hoiem, S.~Chang, M.~Bansal, and H.~Ji, ``Language models with image
  descriptors are strong few-shot video-language learners,'' in \emph{Advances
  in Neural Information Processing Systems}, 2022.

\bibitem{mou2023compressed}
Y.~Mou, X.~Jiang, K.~Xu, T.~Sun, and Z.~Wang, ``Compressed video action
  recognition with dual-stream and dual-modal transformer,'' \emph{IEEE
  Transactions on Circuits and Systems for Video Technology}, 2023.

\bibitem{zheng2023dynamic}
Z.~Zheng, L.~Yang, Y.~Wang, M.~Zhang, L.~He, G.~Huang, and F.~Li, ``Dynamic
  spatial focus for efficient compressed video action recognition,'' \emph{IEEE
  Transactions on Circuits and Systems for Video Technology}, 2023.

\bibitem{ZhaoLG022}
Y.~Zhao, Z.~Li, X.~Guo, and Y.~Lu, ``Alignment-guided temporal attention for
  video action recognition,'' in \emph{Advances in Neural Information
  Processing Systems}, 2022.

\bibitem{slowfast}
C.~Feichtenhofer, H.~Fan, J.~Malik, and K.~He, ``Slowfast networks for video
  recognition,'' in \emph{IEEE International Conference on Computer Vision},
  2019, pp. 6201--6210.

\bibitem{weng2020temporal}
J.~Weng, D.~Luo, Y.~Wang, Y.~Tai, C.~Wang, J.~Li, F.~Huang, X.~Jiang, and
  J.~Yuan, ``Temporal distinct representation learning for action
  recognition,'' in \emph{European Conference on Computer Vision}.\hskip 1em
  plus 0.5em minus 0.4em\relax Springer, 2020, pp. 363--378.

\bibitem{wang2023truncate}
Z.~Wang, J.~Weng, C.~Yuan, and J.~Wang, ``Truncate-split-contrast: a framework
  for learning from mislabeled videos,'' in \emph{Proceedings of the AAAI
  Conference on Artificial Intelligence}, vol.~37, no.~3, 2023, pp. 2751--2758.

\bibitem{dong2022reading}
J.~Dong, Y.~Wang, X.~Chen, X.~Qu, X.~Li, Y.~He, and X.~Wang, ``Reading-strategy
  inspired visual representation learning for text-to-video retrieval,''
  \emph{IEEE Transactions on Circuits and Systems for Video Technology},
  vol.~32, no.~8, pp. 5680--5694, 2022.

\bibitem{ce}
Y.~Liu, S.~Albanie, A.~Nagrani, and A.~Zisserman, ``Use what you have: Video
  retrieval using representations from collaborative experts,'' in
  \emph{British Machine Vision Conference}, 2019, p. 279.

\bibitem{de}
J.~Dong, X.~Li, C.~Xu, S.~Ji, Y.~He, G.~Yang, and X.~Wang, ``Dual encoding for
  zero-example video retrieval,'' in \emph{IEEE Conference on Computer Vision
  and Pattern Recognition}, 2019, pp. 9346--9355.

\bibitem{cao2022locvtp}
M.~Cao, T.~Yang, J.~Weng, C.~Zhang, J.~Wang, and Y.~Zou, ``Locvtp: Video-text
  pre-training for temporal localization,'' in \emph{European Conference on
  Computer Vision}.\hskip 1em plus 0.5em minus 0.4em\relax Springer, 2022, pp.
  38--56.

\bibitem{winner-takes-all}
R.~Zeng, C.~Gan, P.~Chen, W.~Huang, Q.~Wu, and M.~Tan, ``Breaking
  winner-takes-all: Iterative-winners-out networks for weakly supervised
  temporal action localization,'' \emph{IEEE Transactions on Image Processing},
  vol.~28, no.~12, pp. 5797--5808, 2019.

\bibitem{yang2023cross}
J.~Yang, P.~Wei, and N.~Zheng, ``Cross time-frequency transformer for temporal
  action localization,'' \emph{IEEE Transactions on Circuits and Systems for
  Video Technology}, 2023.

\bibitem{sun2022slow}
W.~Sun, R.~Su, Q.~Yu, and D.~Xu, ``Slow motion matters: A slow motion enhanced
  network for weakly supervised temporal action localization,'' \emph{IEEE
  Transactions on Circuits and Systems for Video Technology}, vol.~33, no.~1,
  pp. 354--366, 2022.

\bibitem{lf-vila}
Y.~Sun, H.~Xue, R.~Song, B.~Liu, H.~Yang, and J.~Fu, ``Long-form video-language
  pre-training with multimodal temporal contrastive learning,'' \emph{Advances
  in Neural Information Processing Systems}, vol.~35, pp. 38\,032--38\,045,
  2022.

\bibitem{videograph}
N.~Hussein, E.~Gavves, and A.~W. Smeulders, ``Videograph: Recognizing
  minutes-long human activities in videos,'' \emph{arXiv preprint
  arXiv:1905.05143}, 2019.

\bibitem{svag}
Y.~Wang, G.~Bertasius, T.-H. Oh, A.~Gupta, M.~Hoai, and L.~Torresani,
  ``Supervoxel attention graphs for long-range video modeling,'' in
  \emph{Proceedings of the IEEE/CVF Winter Conference on Applications of
  Computer Vision}, 2021, pp. 155--166.

\bibitem{vis4mer}
M.~M. Islam and G.~Bertasius, ``Long movie clip classification with state-space
  video models,'' in \emph{European Conference on Computer Vision}.\hskip 1em
  plus 0.5em minus 0.4em\relax Springer, 2022, pp. 87--104.

\bibitem{s4nd}
E.~Nguyen, K.~Goel, A.~Gu, G.~Downs, P.~Shah, T.~Dao, S.~Baccus, and C.~R{\'e},
  ``S4nd: Modeling images and videos as multidimensional signals with state
  spaces,'' \emph{Advances in Neural Information Processing Systems}, vol.~35,
  pp. 2846--2861, 2022.

\bibitem{s5}
J.~Wang, W.~Zhu, P.~Wang, X.~Yu, L.~Liu, M.~Omar, and R.~Hamid, ``Selective
  structured state-spaces for long-form video understanding,'' in \emph{IEEE
  Conference on Computer Vision and Pattern Recognition}, 2023, pp. 6387--6397.

\bibitem{s4}
A.~Gu, K.~Goel, and C.~Re, ``Efficiently modeling long sequences with
  structured state spaces,'' in \emph{International Conference on Learning
  Representations}, 2021.

\bibitem{mgsampler}
Y.~Zhi, Z.~Tong, L.~Wang, and G.~Wu, ``Mgsampler: An explainable sampling
  strategy for video action recognition,'' in \emph{IEEE International
  Conference on Computer Vision}, 2021, pp. 1513--1522.

\bibitem{adaframe}
Z.~Wu, C.~Xiong, C.-Y. Ma, R.~Socher, and L.~S. Davis, ``Adaframe: Adaptive
  frame selection for fast video recognition,'' in \emph{IEEE Conference on
  Computer Vision and Pattern Recognition}, 2019, pp. 1278--1287.

\bibitem{barrios2023localizing}
W.~Barrios, M.~Soldan, A.~M. Ceballos-Arroyo, F.~C. Heilbron, and B.~Ghanem,
  ``Localizing moments in long video via multimodal guidance,'' in
  \emph{Proceedings of the IEEE/CVF International Conference on Computer
  Vision}, 2023, pp. 13\,667--13\,678.

\bibitem{mad}
M.~Soldan, A.~Pardo, J.~L. Alc{\'a}zar, F.~Caba, C.~Zhao, S.~Giancola, and
  B.~Ghanem, ``Mad: A scalable dataset for language grounding in videos from
  movie audio descriptions,'' in \emph{Proceedings of the IEEE/CVF Conference
  on Computer Vision and Pattern Recognition}, 2022, pp. 5026--5035.

\bibitem{miech2020end}
A.~Miech, J.-B. Alayrac, L.~Smaira, I.~Laptev, J.~Sivic, and A.~Zisserman,
  ``End-to-end learning of visual representations from uncurated instructional
  videos,'' in \emph{IEEE Conference on Computer Vision and Pattern
  Recognition}, 2020, pp. 9879--9889.

\bibitem{zellers2021merlot}
R.~Zellers, X.~Lu, J.~Hessel, Y.~Yu, J.~S. Park, J.~Cao, A.~Farhadi, and
  Y.~Choi, ``Merlot: Multimodal neural script knowledge models,''
  \emph{Advances in Neural Information Processing Systems}, vol.~34, pp.
  23\,634--23\,651, 2021.

\bibitem{hu2022scaling}
X.~Hu, Z.~Gan, J.~Wang, Z.~Yang, Z.~Liu, Y.~Lu, and L.~Wang, ``Scaling up
  vision-language pre-training for image captioning,'' in \emph{IEEE Conference
  on Computer Vision and Pattern Recognition}, 2022, pp. 17\,980--17\,989.

\bibitem{yang2024gbc}
Z.~Yang, G.~An, Z.~Zheng, S.~Cao, and Q.~Ruan, ``Gbc: Guided alignment and
  adaptive boosting clip bridging vision and language for robust action
  recognition,'' \emph{IEEE Transactions on Circuits and Systems for Video
  Technology}, 2024.

\bibitem{zhang2022unsupervised}
C.~Zhang, T.~Yang, J.~Weng, M.~Cao, J.~Wang, and Y.~Zou, ``Unsupervised
  pre-training for temporal action localization tasks,'' in \emph{IEEE
  Conference on Computer Vision and Pattern Recognition}, 2022, pp.
  14\,031--14\,041.

\bibitem{clip4clip}
H.~Luo, L.~Ji, M.~Zhong, Y.~Chen, W.~Lei, N.~Duan, and T.~Li, ``Clip4clip: An
  empirical study of {CLIP} for end to end video clip retrieval and
  captioning,'' \emph{Neurocomputing}, vol. 508, pp. 293--304, 2022.

\bibitem{actionRecSurvey}
Y.~Kong and Y.~Fu, ``Human action recognition and prediction: A survey,''
  \emph{International Journal of Computer Vision}, vol. 130, no.~5, pp.
  1366--1401, 2022.

\bibitem{pareek2021survey}
P.~Pareek and A.~Thakkar, ``A survey on video-based human action recognition:
  recent updates, datasets, challenges, and applications,'' \emph{Artificial
  Intelligence Review}, vol.~54, pp. 2259--2322, 2021.

\bibitem{li2023lavender}
L.~Li, Z.~Gan, K.~Lin, C.-C. Lin, Z.~Liu, C.~Liu, and L.~Wang, ``Lavender:
  Unifying video-language understanding as masked language modeling,'' in
  \emph{IEEE Conference on Computer Vision and Pattern Recognition}, 2023, pp.
  23\,119--23\,129.

\bibitem{videomae}
Z.~Tong, Y.~Song, J.~Wang, and L.~Wang, ``Videomae: Masked autoencoders are
  data-efficient learners for self-supervised video pre-training,''
  \emph{Advances in Neural Information Processing Systems}, vol.~35, pp.
  10\,078--10\,093, 2022.

\bibitem{videomaev2}
L.~Wang, B.~Huang, Z.~Zhao, Z.~Tong, Y.~He, Y.~Wang, Y.~Wang, and Y.~Qiao,
  ``Videomae v2: Scaling video masked autoencoders with dual masking,'' in
  \emph{IEEE Conference on Computer Vision and Pattern Recognition}, 2023, pp.
  14\,549--14\,560.

\bibitem{violet}
T.-J. Fu, L.~Li, Z.~Gan, K.~Lin, W.~Y. Wang, L.~Wang, and Z.~Liu, ``Violet:
  End-to-end video-language transformers with masked visual-token modeling,''
  \emph{arXiv preprint arXiv:2111.12681}, 2021.

\bibitem{wang2023all}
J.~Wang, Y.~Ge, R.~Yan, Y.~Ge, K.~Q. Lin, S.~Tsutsui, X.~Lin, G.~Cai, J.~Wu,
  Y.~Shan \emph{et~al.}, ``All in one: Exploring unified video-language
  pre-training,'' in \emph{IEEE Conference on Computer Vision and Pattern
  Recognition}, 2023, pp. 6598--6608.

\bibitem{internvideo}
Y.~Wang, K.~Li, Y.~Li, Y.~He, B.~Huang, Z.~Zhao, H.~Zhang, J.~Xu, Y.~Liu,
  Z.~Wang \emph{et~al.}, ``Internvideo: General video foundation models via
  generative and discriminative learning,'' \emph{arXiv preprint
  arXiv:2212.03191}, 2022.

\bibitem{brown2020language}
T.~Brown, B.~Mann, N.~Ryder, M.~Subbiah, J.~D. Kaplan, P.~Dhariwal,
  A.~Neelakantan, P.~Shyam, G.~Sastry, A.~Askell \emph{et~al.}, ``Language
  models are few-shot learners,'' \emph{Advances in Neural Information
  Processing Systems}, vol.~33, pp. 1877--1901, 2020.

\bibitem{glm}
Z.~Du, Y.~Qian, X.~Liu, M.~Ding, J.~Qiu, Z.~Yang, and J.~Tang, ``{GLM:} general
  language model pretraining with autoregressive blank infilling,'' pp.
  320--335, 2022.

\bibitem{llama}
H.~Touvron, T.~Lavril, G.~Izacard, X.~Martinet, M.-A. Lachaux, T.~Lacroix,
  B.~Rozi{\`e}re, N.~Goyal, E.~Hambro, F.~Azhar \emph{et~al.}, ``Llama: Open
  and efficient foundation language models,'' \emph{arXiv preprint
  arXiv:2302.13971}, 2023.

\bibitem{mist}
D.~Gao, L.~Zhou, L.~Ji, L.~Zhu, Y.~Yang, and M.~Z. Shou, ``{MIST} : Multi-modal
  iterative spatial-temporal transformer for long-form video question
  answering,'' in \emph{IEEE Conference on Computer Vision and Pattern
  Recognition}, 2023, pp. 14\,773--14\,783.

\bibitem{jffusion}
Y.~Yu, J.~Kim, and G.~Kim, ``A joint sequence fusion model for video question
  answering and retrieval,'' in \emph{European Conference on Computer Vision},
  vol. 11211, 2018, pp. 487--503.

\bibitem{qa-dataset}
D.~Xu, Z.~Zhao, J.~Xiao, F.~Wu, H.~Zhang, X.~He, and Y.~Zhuang, ``Video
  question answering via gradually refined attention over appearance and
  motion,'' in \emph{Proceedings of ACM international conference on
  Multimedia}, 2017, pp. 1645--1653.

\bibitem{anet-qa}
Z.~Yu, D.~Xu, J.~Yu, T.~Yu, Z.~Zhao, Y.~Zhuang, and D.~Tao, ``Activitynet-qa:
  {A} dataset for understanding complex web videos via question answering,'' in
  \emph{AAAI Conference on Artificial Intelligence}, 2019, pp. 9127--9134.

\bibitem{LinLL0G0LW22}
K.~Lin, L.~Li, C.~Lin, F.~Ahmed, Z.~Gan, Z.~Liu, Y.~Lu, and L.~Wang,
  ``Swinbert: End-to-end transformers with sparse attention for video
  captioning,'' in \emph{IEEE Conference on Computer Vision and Pattern
  Recognition}, 2022, pp. 17\,928--17\,937.

\bibitem{vid2seq}
A.~Yang, A.~Nagrani, P.~H. Seo, A.~Miech, J.~Pont-Tuset, I.~Laptev, J.~Sivic,
  and C.~Schmid, ``Vid2seq: Large-scale pretraining of a visual language model
  for dense video captioning,'' in \emph{IEEE Conference on Computer Vision and
  Pattern Recognition}, 2023, pp. 10\,714--10\,726.

\bibitem{anetcap}
R.~Krishna, K.~Hata, F.~Ren, L.~Fei{-}Fei, and J.~C. Niebles,
  ``Dense-captioning events in videos,'' in \emph{IEEE International Conference
  on Computer Vision}, 2017, pp. 706--715.

\bibitem{blip2}
J.~Li, D.~Li, S.~Savarese, and S.~Hoi, ``{BLIP}-2: Bootstrapping language-image
  pre-training with frozen image encoders and large language models,'' in
  \emph{International Conference on Machine Learning}, vol. 202, 23--29 Jul
  2023, pp. 19\,730--19\,742.

\bibitem{zhu2023minigpt}
D.~Zhu, J.~Chen, X.~Shen, X.~Li, and M.~Elhoseiny, ``Minigpt-4: Enhancing
  vision-language understanding with advanced large language models,''
  \emph{arXiv preprint arXiv:2304.10592}, 2023.

\bibitem{internvid}
Y.~Wang, Y.~He, Y.~Li, K.~Li, J.~Yu, X.~Ma, X.~Chen, Y.~Wang, P.~Luo, Z.~Liu
  \emph{et~al.}, ``Internvid: A large-scale video-text dataset for multimodal
  understanding and generation,'' \emph{arXiv preprint arXiv:2307.06942}, 2023.

\bibitem{scenedetect}
``Pyscenedetect,'' https://github.com/Breakthrough/PySceneDetect.

\bibitem{decord}
``Decord: An efficient video loader for deep learning,''
  https://github.com/dmlc/decord.

\bibitem{clip}
A.~Radford, J.~W. Kim, C.~Hallacy, A.~Ramesh, G.~Goh, S.~Agarwal, G.~Sastry,
  A.~Askell, P.~Mishkin, J.~Clark \emph{et~al.}, ``Learning transferable visual
  models from natural language supervision,'' in \emph{International Conference
  on Machine Learning}, 2021, pp. 8748--8763.

\bibitem{dino}
H.~Zhang, F.~Li, S.~Liu, L.~Zhang, H.~Su, J.~Zhu, L.~M. Ni, and H.~Shum,
  ``{DINO:} {DETR} with improved denoising anchor boxes for end-to-end object
  detection,'' in \emph{International Conference on Learning Representations},
  2023.

\bibitem{sentencebert}
N.~Reimers and I.~Gurevych, ``Sentence-bert: Sentence embeddings using siamese
  bert-networks,'' in \emph{Proceedings of Conference on Empirical Methods in
  Natural Language Processing and International Joint Conference on Natural
  Language Processing}, 2019, pp. 3980--3990.

\bibitem{prvr}
J.~Dong, X.~Chen, M.~Zhang, X.~Yang, S.~Chen, X.~Li, and X.~Wang, ``Partially
  relevant video retrieval,'' in \emph{Proceedings of ACM International
  Conference on Multimedia}, 2022, pp. 246--257.

\bibitem{jsg-prvr}
Z.~Chen, X.~Jiang, X.~Xu, Z.~Cao, Y.~Mo, and H.~T. Shen, ``Joint searching and
  grounding: Multi-granularity video content retrieval,'' in \emph{Proceedings
  of ACM International Conference on Multimedia}, 2023, pp. 975--983.

\bibitem{hit}
S.~Liu, H.~Fan, S.~Qian, Y.~Chen, W.~Ding, and Z.~Wang, ``Hit: Hierarchical
  transformer with momentum contrast for video-text retrieval,'' in \emph{IEEE
  International Conference on Computer Vision}, 2021, pp. 11\,895--11\,905.

\bibitem{Gabeur0AS20}
V.~Gabeur, C.~Sun, K.~Alahari, and C.~Schmid, ``Multi-modal transformer for
  video retrieval,'' in \emph{European Conference on Computer Vision}, vol.
  12349, 2020, pp. 214--229.

\bibitem{ZhangHS18}
B.~Zhang, H.~Hu, and F.~Sha, ``Cross-modal and hierarchical modeling of video
  and text,'' in \emph{European Conference on Computer Vision}, vol. 11217,
  2018, pp. 385--401.

\bibitem{eclipse}
Y.~Lin, J.~Lei, M.~Bansal, and G.~Bertasius, ``Eclipse: Efficient long-range
  video retrieval using sight and sound,'' in \emph{European Conference on
  Computer Vision}, vol. 13694, 2022, pp. 413--430.

\bibitem{charades-sta}
J.~Gao, C.~Sun, Z.~Yang, and R.~Nevatia, ``{TALL:} temporal activity
  localization via language query,'' in \emph{IEEE International Conference on
  Computer Vision}, 2017, pp. 5277--5285.

\bibitem{charades}
G.~A. Sigurdsson, G.~Varol, X.~Wang, A.~Farhadi, I.~Laptev, and A.~Gupta,
  ``Hollywood in homes: Crowdsourcing data collection for activity
  understanding,'' in \emph{European Conference on Computer Vision}, 2016, pp.
  510--526.

\bibitem{msrvtt}
J.~Xu, T.~Mei, T.~Yao, and Y.~Rui, ``{MSR-VTT:} {A} large video description
  dataset for bridging video and language,'' in \emph{IEEE Conference on
  Computer Vision and Pattern Recognition}, 2016, pp. 5288--5296.

\bibitem{ques-gener}
M.~Heilman and N.~A. Smith, ``Question generation via overgenerating
  transformations and ranking,'' \emph{DTIC Document}, 2009.

\bibitem{ego_schema}
K.~Mangalam, R.~Akshulakov, and J.~Malik, ``Egoschema: {A} diagnostic benchmark
  for very long-form video language understanding,'' in \emph{Advances in
  Neural Information Processing Systems}, A.~Oh, T.~Naumann, A.~Globerson,
  K.~Saenko, M.~Hardt, and S.~Levine, Eds., 2023.

\bibitem{ego4d}
K.~Grauman, A.~Westbury, E.~Byrne, Z.~Chavis, A.~Furnari, R.~Girdhar,
  J.~Hamburger, H.~Jiang, M.~Liu, X.~Liu, M.~Martin, T.~Nagarajan,
  I.~Radosavovic, S.~K. Ramakrishnan, F.~Ryan, J.~Sharma, M.~Wray, M.~Xu, E.~Z.
  Xu, C.~Zhao, S.~Bansal, D.~Batra, V.~Cartillier, S.~Crane, T.~Do, M.~Doulaty,
  A.~Erapalli, C.~Feichtenhofer, A.~Fragomeni, Q.~Fu, A.~Gebreselasie,
  C.~Gonz{\'{a}}lez, J.~Hillis, X.~Huang, Y.~Huang, W.~Jia, W.~Khoo,
  J.~Kol{\'{a}}r, S.~Kottur, A.~Kumar, F.~Landini, C.~Li, Y.~Li, Z.~Li,
  K.~Mangalam, R.~Modhugu, J.~Munro, T.~Murrell, T.~Nishiyasu, W.~Price, P.~R.
  Puentes, M.~Ramazanova, L.~Sari, K.~Somasundaram, A.~Southerland, Y.~Sugano,
  R.~Tao, M.~Vo, Y.~Wang, X.~Wu, T.~Yagi, Z.~Zhao, Y.~Zhu, P.~Arbel{\'{a}}ez,
  D.~Crandall, D.~Damen, G.~M. Farinella, C.~Fuegen, B.~Ghanem, V.~K. Ithapu,
  C.~V. Jawahar, H.~Joo, K.~Kitani, H.~Li, R.~A. Newcombe, A.~Oliva, H.~S.
  Park, J.~M. Rehg, Y.~Sato, J.~Shi, M.~Z. Shou, A.~Torralba, L.~Torresani,
  M.~Yan, and J.~Malik, ``Ego4d: Around the world in 3, 000 hours of egocentric
  video,'' in \emph{IEEE Conference on Computer Vision and Pattern
  Recognition}.\hskip 1em plus 0.5em minus 0.4em\relax {IEEE}, 2022, pp.
  18\,973--18\,990.

\bibitem{nextqa}
J.~Xiao, X.~Shang, A.~Yao, and T.~Chua, ``Next-qa: Next phase of
  question-answering to explaining temporal actions,'' in \emph{IEEE Conference
  on Computer Vision and Pattern Recognition}, 2021, pp. 9777--9786.

\bibitem{kinetics}
J.~Carreira and A.~Zisserman, ``Quo vadis, action recognition? a new model and
  the kinetics dataset,'' in \emph{IEEE Conference on Computer Vision and
  Pattern Recognition}, 2017, pp. 6299--6308.

\bibitem{vicuna}
L.~Zheng, W.-L. Chiang, Y.~Sheng, S.~Zhuang, Z.~Wu, Y.~Zhuang, Z.~Lin, Z.~Li,
  D.~Li, E.~Xing \emph{et~al.}, ``Judging llm-as-a-judge with mt-bench and
  chatbot arena,'' \emph{arXiv preprint arXiv:2306.05685}, 2023.

\bibitem{sharegpt}
``Sharegpt: Share your wildest chatgpt conversations with one click,'' 2023,
  https://github.com/domeccleston/sharegpt.

\bibitem{angle}
X.~Li and J.~Li, ``Angle-optimized text embeddings,'' \emph{arXiv preprint
  arXiv:2309.12871}, 2023.

\bibitem{relocnet}
H.~Zhang, A.~Sun, W.~Jing, G.~Nan, L.~Zhen, J.~T. Zhou, and R.~S.~M. Goh,
  ``Video corpus moment retrieval with contrastive learning,'' in
  \emph{Proceedings of International ACM SIGIR Conference on Research and
  Development in Information Retrieval}, 2021, pp. 685--695.

\bibitem{xml}
J.~Lei, L.~Yu, T.~L. Berg, and M.~Bansal, ``{TVR:} {A} large-scale dataset for
  video-subtitle moment retrieval,'' in \emph{European Conference on Computer
  Vision}, vol. 12366, 2020, pp. 447--463.

\bibitem{DLDKD}
J.~Dong, M.~Zhang, Z.~Zhang, X.~Chen, D.~Liu, X.~Qu, X.~Wang, and B.~Liu,
  ``Dual learning with dynamic knowledge distillation for partially relevant
  video retrieval,'' in \emph{IEEE International Conference on Computer
  Vision}, 2023.

\bibitem{videollava}
B.~Lin, B.~Zhu, Y.~Ye, M.~Ning, P.~Jin, and L.~Yuan, ``Video-llava: Learning
  united visual representation by alignment before projection,'' \emph{arXiv
  preprint arXiv:2311.10122}, 2023.

\bibitem{justask}
A.~Yang, A.~Miech, J.~Sivic, I.~Laptev, and C.~Schmid, ``Just ask: Learning to
  answer questions from millions of narrated videos,'' in \emph{IEEE
  International Conference on Computer Vision}, 2021, pp. 1666--1677.

\bibitem{merlotreserve}
R.~Zellers, J.~Lu, X.~Lu, Y.~Yu, Y.~Zhao, M.~Salehi, A.~Kusupati, J.~Hessel,
  A.~Farhadi, and Y.~Choi, ``{MERLOT} {RESERVE:} neural script knowledge
  through vision and language and sound,'' in \emph{IEEE Conference on Computer
  Vision and Pattern Recognition}, 2022, pp. 16\,354--16\,366.

\bibitem{frozenbilm}
A.~Yang, A.~Miech, J.~Sivic, I.~Laptev, and C.~Schmid, ``Zero-shot video
  question answering via frozen bidirectional language models,'' in
  \emph{Advances in Neural Information Processing Systems}, 2022.

\bibitem{hitea}
Q.~Ye, G.~Xu, M.~Yan, H.~Xu, Q.~Qian, J.~Zhang, and F.~Huang, ``Hitea:
  Hierarchical temporal-aware video-language pre-training,'' in \emph{IEEE
  International Conference on Computer Vision}, 2023, pp. 15\,405--15\,416.

\bibitem{yu2024self}
S.~Yu, J.~Cho, P.~Yadav, and M.~Bansal, ``Self-chained image-language model for
  video localization and question answering,'' \emph{Advances in Neural
  Information Processing Systems}, vol.~36, 2024.

\bibitem{balazevic2024Memory}
I.~Balazevic, Y.~Shi, P.~Papalampidi, R.~Chaabouni, S.~Koppula, and O.~J.
  Henaff, ``Memory consolidation enables long-context video understanding,'' in
  \emph{International Conference on Machine Learning}, vol. 235, 2024, pp.
  2527--2542.

\bibitem{ye2023mplug}
Q.~Ye, H.~Xu, G.~Xu, J.~Ye, M.~Yan, Y.~Zhou, J.~Wang, A.~Hu, P.~Shi, Y.~Shi
  \emph{et~al.}, ``mplug-owl: Modularization empowers large language models
  with multimodality,'' \emph{arXiv preprint arXiv:2304.14178}, 2023.

\bibitem{llovi}
C.~Zhang, T.~Lu, M.~M. Islam, Z.~Wang, S.~Yu, M.~Bansal, and G.~Bertasius, ``A
  simple {LLM} framework for long-range video question-answering,''
  \emph{CoRR}, vol. abs/2312.17235, 2023.

\bibitem{buch2022revisiting}
S.~Buch, C.~Eyzaguirre, A.~Gaidon, J.~Wu, L.~Fei-Fei, and J.~C. Niebles,
  ``Revisiting the" video" in video-language understanding,'' in \emph{IEEE
  Conference on Computer Vision and Pattern Recognition}, 2022, pp. 2917--2927.

\bibitem{xiao2022vgt}
J.~Xiao, P.~Zhou, T.-S. Chua, and S.~Yan, ``Video graph transformer for video
  question answering,'' in \emph{European Conference on Computer Vision}.\hskip
  1em plus 0.5em minus 0.4em\relax Springer, 2022, pp. 39--58.

\bibitem{momeni2023verbs}
L.~Momeni, M.~Caron, A.~Nagrani, A.~Zisserman, and C.~Schmid, ``Verbs in
  action: Improving verb understanding in video-language models,'' in
  \emph{IEEE International Conference on Computer Vision}, 2023, pp.
  15\,579--15\,591.

\bibitem{alayrac2022flamingo}
J.-B. Alayrac, J.~Donahue, P.~Luc, A.~Miech, I.~Barr, Y.~Hasson, K.~Lenc,
  A.~Mensch, K.~Millican, M.~Reynolds \emph{et~al.}, ``Flamingo: a visual
  language model for few-shot learning,'' \emph{Advances in Neural Information
  Processing Systems}, vol.~35, pp. 23\,716--23\,736, 2022.

\bibitem{wang2024videoagent}
X.~Wang, Y.~Zhang, O.~Zohar, and S.~Yeung-Levy, ``Videoagent: Long-form video
  understanding with large language model as agent,'' \emph{arXiv preprint
  arXiv:2403.10517}, 2024.

\bibitem{alpro}
D.~Li, J.~Li, H.~Li, J.~C. Niebles, and S.~C.~H. Hoi, ``Align and prompt:
  Video-and-language pre-training with entity prompts,'' in \emph{IEEE
  Conference on Computer Vision and Pattern Recognition}, 2022, pp. 4943--4953.

\bibitem{singularity}
J.~Lei, T.~L. Berg, and M.~Bansal, ``Revealing single frame bias for
  video-and-language learning,'' in \emph{Proceedings of Annual Meeting of the
  Association for Computational Linguistics}, 2023, pp. 487--507.

\end{thebibliography}
